%% file: main.tex
\documentclass[sigconf]{acmart}
\usepackage{colortbl}
\usepackage{xcolor}
\usepackage{multirow}
\usepackage{graphicx}
\usepackage{wrapfig}
\usepackage{pifont}
\usepackage{algorithm}
\usepackage{algpseudocode}

\AtBeginDocument{%
  }

\copyrightyear{2025}
\acmYear{2025}
\setcopyright{acmlicensed}\acmConference[KDD '25]{Proceedings of the 31st ACM SIGKDD Conference on Knowledge Discovery and Data Mining V.2}{August 3--7, 2025}{Toronto, ON, Canada}
\acmBooktitle{Proceedings of the 31st ACM SIGKDD Conference on Knowledge Discovery and Data Mining V.2 (KDD '25), August 3--7, 2025, Toronto, ON, Canada}
\acmDOI{10.1145/3711896.3737248}
\acmISBN{979-8-4007-1454-2/2025/08}

\begin{document}

\title{Mini-Game Lifetime Value Prediction in WeChat}

\author{Aochuan Chen}
\authornote{Authors contributed equally to this research.}
\email{achen149@connect.hkust-gz.edu.cn}
\orcid{0009-0002-2300-1498}
\affiliation{%
  \institution{The Hong Kong University of Science and Technology (Guangzhou)}
  \city{Guangzhou}
  \state{Guangdong}
  \country{China}
}

\author{Yifan Niu}
\authornotemark[1]
\email{yniu669@connect.hkust-gz.edu.cn}
\orcid{0009-0004-1985-7044}
\affiliation{%
  \institution{The Hong Kong University of Science and Technology (Guangzhou)}
  \city{Guangzhou}
  \state{Guangdong}
  \country{China}
}

\author{Ziqi Gao}
\authornotemark[1]
\email{zgaoat@connect.ust.hk}
\orcid{0000-0002-7417-3620}
\affiliation{%
  \institution{The Hong Kong University of Science and Technology (Guangzhou)}
  \city{Guangzhou}
  \state{Guangdong}
  \country{China}
}

\author{Yujie Sun}
\email{sebastisun@tencent.com}
\orcid{0009-0000-6660-776X}
\affiliation{%
  \institution{Tencent Inc.}
  \city{Shenzhen}
  \state{Guangdong}
  \country{China}
}

\author{Shoujun Liu}
\email{aldenliu@tencent.com}
\orcid{0009-0003-0676-4407}
\affiliation{%
  \institution{Tencent Inc.}
  \city{Shenzhen}
  \state{Guangdong}
  \country{China}
}

\author{Gong Chen}
\email{natchen@tencent.com}
\orcid{0009-0004-1124-2571}
\affiliation{%
  \institution{Tencent Inc.}
  \city{Shenzhen}
  \state{Guangdong}
  \country{China}
}

\author{Yang Liu}
\email{yliukj@connect.ust.hk}
\orcid{0000-0002-2633-512X}
\affiliation{%
  \institution{The Hong Kong University of Science and Technology (Guangzhou)}
  \city{Guangzhou}
  \state{Guangdong}
  \country{China}
}

\author{Jia Li}
\authornote{Corresponding Author.}
\email{jialee@hkust-gz.edu.cn}
\orcid{0000-0002-6362-4385}
\affiliation{%
  \institution{The Hong Kong University of Science and Technology (Guangzhou)}
  \city{Guangzhou}
  \state{Guangdong}
  \country{China}
}

\renewcommand{\shortauthors}{Aochuan Chen et al.}
\newcommand{\ourmeth}{GRePO-LTV}




\input{tex/abstract}
\maketitle
\input{tex/introduction}

\input{tex/preliminary}

\input{tex/method}

\input{tex/experiment}
\input{tex/lessons}
\input{tex/related_work}
\input{tex/conclusion}
\input{tex/acknowledgment}

\bibliographystyle{ACM-Reference-Format}
\bibliography{references}

\appendix
\input{tex/appendix}

\end{document}

%% file: tex/abstract.tex
\begin{abstract}
The LifeTime Value (LTV) prediction, which endeavors to forecast the cumulative purchase contribution of a user to a particular item, remains a vital challenge that advertisers are keen to resolve. A precise LTV prediction system enhances the alignment of user interests with meticulously designed advertisements, thereby generating substantial profits for advertisers. Nonetheless, this issue is complicated by the paucity of data typically observed in real-world advertising scenarios. The purchase rate among registered users is often as critically low as 0.1\%, resulting in a dataset where the majority of users make only several purchases. Consequently, there is insufficient supervisory signal for effectively training the LTV prediction model. 
An additional challenge emerges from the interdependencies among tasks with high correlation. It is a common practice to estimate a user's contribution to a game over a specified temporal interval. Varying the lengths of these intervals corresponds to distinct predictive tasks, which are highly correlated. For instance, predictions over a 7-day period are heavily reliant on forecasts made over a 3-day period, where exceptional cases can adversely affect the accuracy of both tasks.
In order to comprehensively address the aforementioned challenges, we introduce an innovative framework denoted as \textit{Graph-Represented Pareto-Optimal LifeTime Value prediction (\ourmeth)}. Graph representation learning is initially employed to address the issue of data scarcity. Subsequently, Pareto-Optimization is utilized to manage the interdependence of prediction tasks. Our method is evaluated using a proprietary offline mini-game recommendation dataset in conjunction with an online A/B test. The implementation of our method results in a significant enhancement within the offline dataset. Moreover, the A/B test demonstrates encouraging outcomes, increasing average Gross Merchandise Value (GMV) by 8.4\%.
\end{abstract}

%% file: tex/introduction.tex
\section{Introduction}

In the competitive gaming industry, predicting user Lifetime Value (LTV) is crucial for optimizing acquisition, engagement, monetization, and retention strategies. LTV prediction enables data-driven resource allocation to boost long-term profitability, aids efficient user acquisition by identifying high-value users, enhances engagement with personalized experiences, guides optimal game design and virtual goods pricing, supports profitable LiveOps planning, and triggers proactive retention for high-value users at risk of churning. Companies using advanced data science to predict LTV gain a significant competitive edge in a saturated market.

Various methods for predicting LTV have been proposed, ranging from traditional approaches like RFM \cite{fader2005rfm} and Markov Chain models \cite{norris1998markov} to machine learning techniques such as random forests \cite{lariviere2005predicting} and gradient boosting \cite{singh2018customer}. Recently, deep learning architectures have pushed the state-of-the-art by capturing complex patterns in high-dimensional data \cite{drachen2018or, zhang2023out, zhou2024cross}.

Despite significant advancements in LTV prediction methods, a fundamental challenge persists that hinders accurate forecasting. This difficulty arises from the nature of the LTV prediction task itself, which sits at the end of the advertising conversion funnel, typically represented as ``exposure -> click -> register -> purchase''. As we progress deeper into the conversion funnel, the amount of available data diminishes significantly at each stage. At the purchase stage, we face a severe shortage of data, posing substantial obstacles to accurate LTV prediction due to limited sample size, high variance, and noise in the sparse data points. 

The time domain presents another challenge in LTV prediction, as multi-period prediction is employed to compensate for limited data. Just like time-series prediction, excelling in both short-term and long-term predictions is inherently difficult \cite{zhang2024elastst}, as the dynamics governing these time scales often differ significantly. Optimizing a model for both can be detrimental, as features relevant for one may not be informative or may mislead the other. This trade-off between short-term and long-term accuracy hinders the development of a unified model across all time horizons.

To address the aforementioned challenges, we propose \textbf{G}raph-\textbf{Re}presented \textbf{P}areto-\textbf{O}ptimal LifeTime Value prediction (\ourmeth). Our method adopts graph representation learning to alleviate the paucity of purchase records and build effective user and game embeddings. By learning graph-based representations that capture the complex interactions between users and games, we can better encode the collaborative signal \cite{wang2019neural} and utilize the limited available data to improve LTV prediction accuracy. Furthermore, our method employs Pareto optimization to avoid task contradiction between value predictions of different time horizons. This multi-objective optimization approach allows us to find a set of Pareto-optimal solutions that balance the trade-offs between short-term and long-term prediction accuracy. By considering the Pareto front, we can select a model that achieves strong performance across all time horizons without compromising on either end of the spectrum.

Our contributions are summarized below:

\begin{enumerate}
    \item Graph representation learning leverages user-game interactions to alleviate data sparsity and build effective embeddings, enhancing LTV prediction accuracy.
    \item Pareto optimization addresses task contradiction between short-term and long-term value predictions, balancing trade-offs to achieve strong performance across all time horizons.
    \item The proposed method, \ourmeth, demonstrates noticeable performance boosts in both offline experiments and online A/B testing compared to existing approaches.
\end{enumerate}

%% file: tex/preliminary.tex
\section{Preliminary}
\label{sec:preliminary}

\subsection{Problem Definition}
\label{sec:problem_definition}

In formal terms, there exists a set of users $\mathcal{U} = \{u_p\}_{p=1}^{P}$, accompanied by a corresponding feature set $\mathcal{X}_u = \{x_{u_p}\}_{p=1}^P$ and a sequence of behavioral patterns $\mathcal{B}_u = \{b_{u_p}\}_{p=1}^P$. Additionally, there is a set of items $\mathcal{I} = \{i_q\}_{q=1}^Q$, endowed with its own feature set $\mathcal{X}_i = \{x_{i_q}\}_{q=1}^Q$. Each $b_{u_p}$ is constituted by an array of items that have previously been interacted with by the user $u_p$.

\begin{definition}[$t$-Value Prediction]
We define $t$-Value as the value a user contributes within a period lasting $t$ days. For instance, the value contributed by a user in the initial $7$ days following registration is referred to as $7$-Value. Consequently, a user's LTV can be characterized as $\infty$-Value. The LTV problem can be deconstructed into sub-problems at specified time intervals. These sub-problems provide more robust supervisory signals to improve the model's generalization capability. Mathmatically, the $t$-Value prediction problem is defined as
\begin{equation*}
    \hat{y}_{t-\mathrm{value}}^{u_p,i_q} = f(u_p,x_{u_p},b_{u_p},i_q,x_{i_q} | \Theta_{t-\mathrm{value}}) \in [0,+\infty).
\end{equation*}
\end{definition}

\subsection{Advertising Conversion Funnel}
\label{sec:ad_conversion_funnel}

\begin{figure}[htb]
    \centering
    \includegraphics[width=\linewidth]{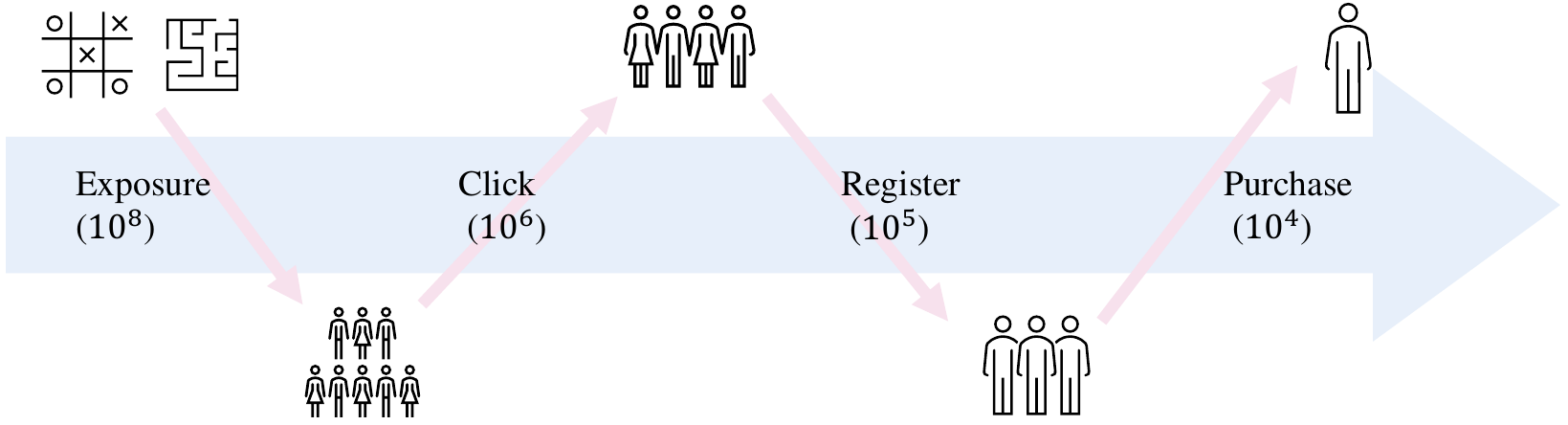}
    \caption{An illustration for the advertising conversion funnel, which shows the user journey from initial exposure to purchase. As users move through each stage, fewer reach the final purchase, resulting in sparse data for purchased users.}
    \label{fig:conversion_funnel}
\end{figure}

The advertising conversion funnel has four stages: exposure (potential customer sees the ad), click (customer clicks on the ad), register (customer registers an account), and purchase (customer buys the product). The conversion rate between each stage is very low, often in the single digits. For example, if an ad reaches 100 million people, a 2\% click-through rate means 2 million website visitors. If 25\% of visitors register, the lead pool shrinks to 0.5 million. With a 1\% conversion rate from lead to customer, the campaign would yield just 500 purchases. As a result, the number of customers who complete the final purchase is a tiny fraction of those initially exposed to the advertisement. We provide a detailed conversion funnel figure for mini-game advertising on the WeChat platform in Figure \ref{fig:conversion_funnel}.

\subsection{Time Horizons}
\label{sec:time_horizons}

\begin{figure}[htb]
    \centering
    \includegraphics[width=0.8\linewidth]{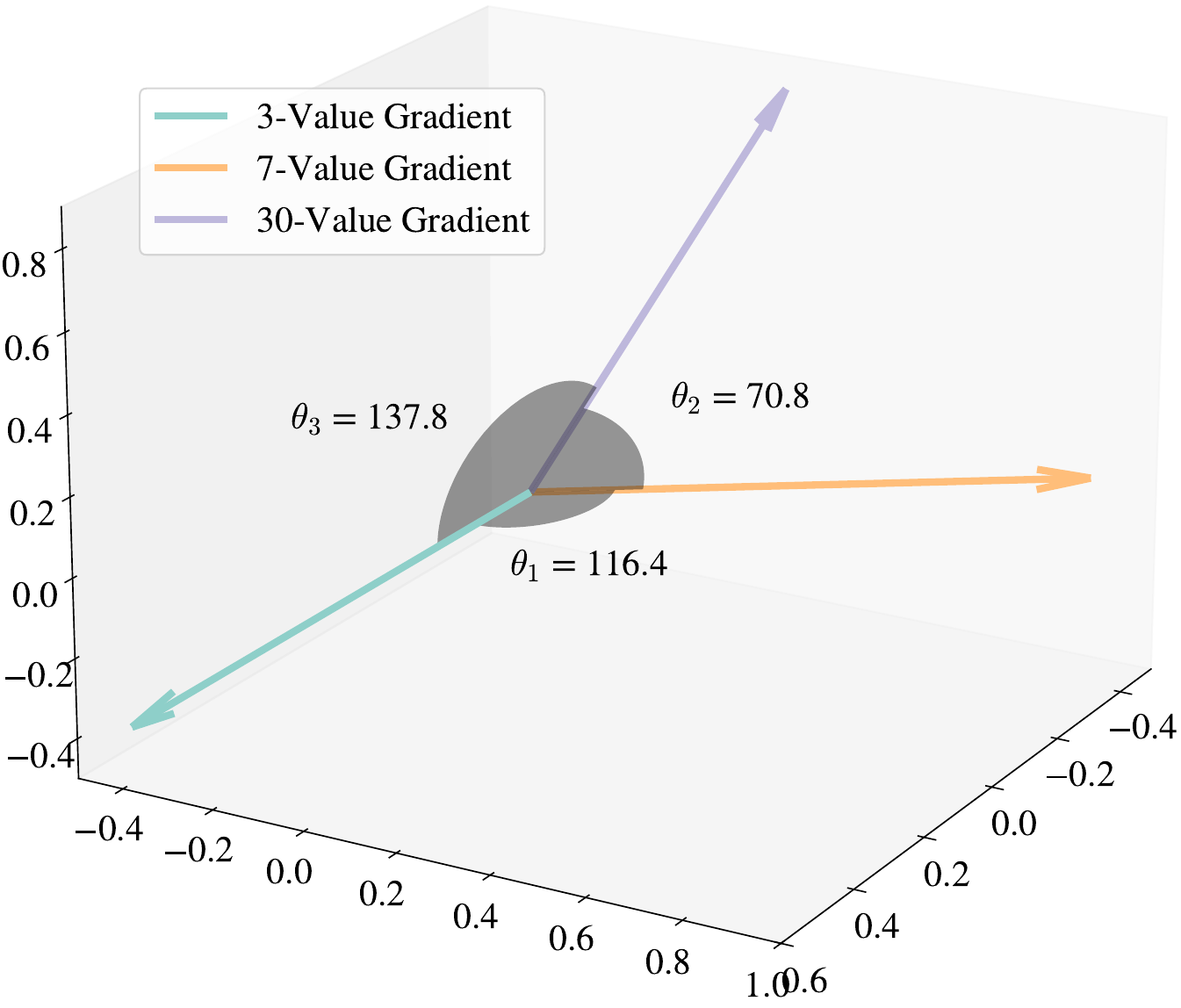}
    \caption{Gradient conflict in multi-period prediction. We observe that the gradient of 3-Value prediction is always in contradiction with the other two.}
    \label{fig:pareto_gra}
\end{figure}%

Advertising companies predict user value over specific horizons (\textit{e.g.,} 3 days, 7 days, 30 days) \citep{10.1145/3580305.3599871}. 
However, tasks targeting different horizons have complex interconnections \citep{li2022billion}. For example, short-term value predictions should be strictly smaller than long-term predictions for the same user. Previous work imposed this restriction by modifying the network structure \citep{li2022billion}. We instead ask: What if these tasks are co-trained on the same network? Would they collaborate to enhance network expressiveness or conflict with each other?
To investigate this, we observed gradients of different tasks trained on the same neural network. 
In multi-period prediction, the gradient conflict (angles $\theta > \frac{\pi}{2}$) phenomenon is commonly observed across different periods. To keep the gradient angle, we use the orthogonal matrix and Linear Transformation to map the gradients of three tasks to a three-dimensional space, as illustrated in Figure~\ref{fig:pareto_gra}. When the model descends along the gradient of one period, the objective function of the conflicting period will increase. Therefore, it often results in imbalanced performance and low precision for certain periods~\cite{deb2016multi}. To meet practical industrial needs, it is crucial to develop a model that is balanced, rather than skewed, and fulfills the prediction accuracy requirements for each period.  We should strike a balance between short-term and long-term accuracy, developing a unified model that performs well across all time horizons.

%% file: tex/method.tex
\section{Method}

\begin{figure*}[t]
\centering
\includegraphics[width=1.0\textwidth]{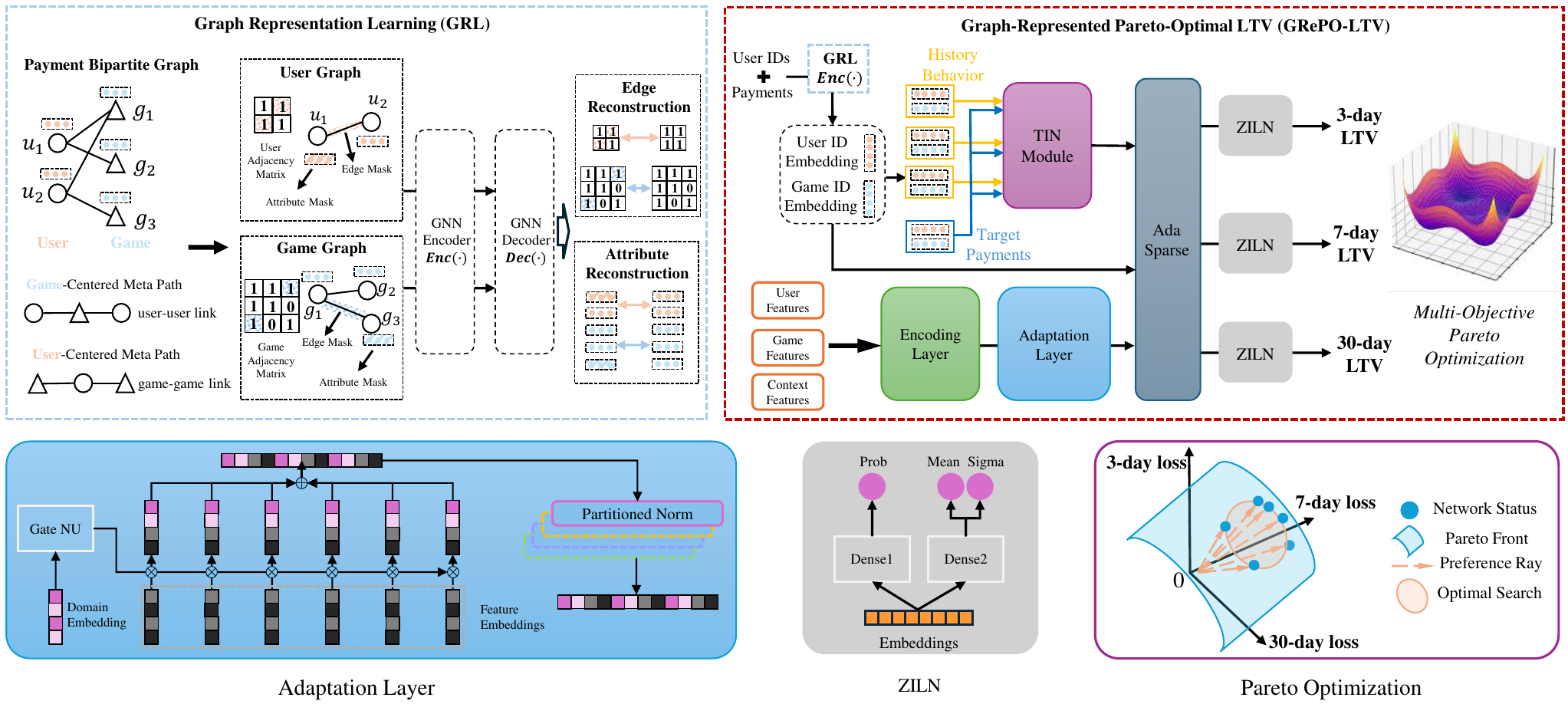} 
\caption{An overview of \ourmeth. 
User and game embeddings are first obtained through GRL. Subsequently, GRePO-LTV is constructed using an encoding layer, an adaptation layer, a Temporal Interaction Network (TIN) module, and tower layers. The model is optimized with Pareto optimization technique to effectively balance multiple objectives of different time horizons.
}
\label{fig:main}
\end{figure*}

Figure \ref{fig:main} provides an overview of Graph Representation Learning (GRL) and \ourmeth.

\subsection{Graph Representation Learning}
\label{sec:graph_learn}
Obtaining substantial supervisory signals for determining the LTV of individual users is a significant challenge in many practical applications. This difficulty arises from the fact that most users exhibit sparse purchasing behavior, engaging in transactions only infrequently \cite{chen2025graph}. One promising approach to mitigate this limitation is the utilization of graph representation learning techniques to capture collaborative signals \cite{he2020lightgcn}. By leveraging the inherent structure of user-item interaction graphs, these methods enable the learning of robust user and item embeddings even in scenarios characterized by limited availability of direct behavioral data for specific users or items. Recent research has demonstrated the effectiveness of graph representation learning in enhancing the quality of embeddings \cite{sun2019multi, li2025g}. This line of work highlights the potential of graph-based approaches in addressing the challenges posed by sparse user activity data and facilitating more accurate estimation of user LTV.

Building on the methodology described in \cite{tian2023heterogeneous}, homogeneous graphs are first constructed by identifying `meta paths', which include user-game-user and game-user-game interaction patterns. Specifically, a user graph is generated where an edge exists between two users if they have interacted with a common game. The weight of this edge is proportional to the count of games they have both interacted with. Correspondingly, a game graph is formed by establishing an edge between two games if they have been interacted with by the same user; the edge strength is determined by the number of such common users. For the purpose of unsupervised learning, a masking technique is employed, wherein a random subset of edges and node attributes is concealed. The objective of the training process is to reconstruct these masked elements, thereby enabling the model to learn robust representations of nodes.

Our loss function is composed of two fundamental parts, the edge reconstruction and the attribute reconstruction. In math, the edge reconstruction can be described as
\begin{align*}
    l_{\mathrm{e-recon}} = \frac{1}{\|\Phi\|}\sum_{\phi\in\Phi} \frac{1}{\|\mathcal{A}^{\phi}\|}\sum_{v\in\mathcal{V}^{\phi}} (1-\frac{\mathcal{A}^{\phi}_{v}\cdot \hat{\mathcal{A}}^{\phi}_{v}}{\|\mathcal{A}^{\phi}_{v}\|\times\|\hat{\mathcal{A}}^{\phi}_{v}\|})^{\xi_e},
\end{align*}
where $\Phi$ denotes the set of meta paths (in our setting, $\|\Phi\|=2$), $\mathcal{A}$ and $\hat{\mathcal{A}}$ represent the adjacency matrix and its reconstructed version, respectively. $\mathcal{V}^{\phi}$ is the set of nodes for meta path $\phi$, and $\xi_e$ is a hyperparameter used to scale the reconstruction loss. Similarly, the attribute reconstruction loss is characterized as
\begin{align*}
    l_{\mathrm{a-recon}} = \frac{1}{\|\mathcal{V}_{\mathrm{msk}}\|}\sum_{v\in\mathcal{V}_{\mathrm{msk}}}(1-\frac{x_v\cdot \hat{x}_{v}}{\|x_v\|\times \|\hat{x}_{v}\|})^{\xi_a},
\end{align*}
where $\mathcal{V}_{\mathrm{msk}}$ represents the set of masked nodes, $x_v$ and $\hat{x}_v$ denote the attribute vectors of node $v$ and their reconstructed versions after masking, respectively, and $\xi_a$ is a scaling coefficient for the attribute reconstruction loss. We combine these two losses using a hyperparameter $\zeta$:
\begin{align*}
    \mathcal{L}_{\mathrm{GRL}} = l_{\mathrm{a-recon}} + \zeta l_{\mathrm{e-recon}}.
\end{align*}

\subsection{Backbone of \ourmeth}
\label{sec:backbone}

\ourmeth~consists of four key components: the encoding layer, adaptation layer, TIN module, and tower layer.

\paragraph{\textbf{Encoding Layer}}
The feature processing pipeline begins with categorical user and item attributes, which undergo embedding transformation through embedding lookup tables. This process maps each categorical value to a dense vector representation in a learnable embedding space.

Each feature is organized into distinct fields, preserving the semantic structure of different attribute types. For instance, user demographics, behavioral features, and item characteristics are treated as separate fields. This field-based organization enables the model to learn field-specific interaction patterns. The Field-weighted Factorization Machine (FwFM) \cite{pan2018field} then models cross-field interactions by introducing field pair weights. 

\paragraph{\textbf{Adaptation Layer}} 
The adaptation layer combines two domain adaptation mechanisms: EPNet \cite{chang2023pepnet} and Partitioned Norm \cite{sheng2021one}. EPNet introduces domain-specific feature modulation through a gate neural unit (Gate NU), while Partitioned Norm handles domain-specific feature normalization.

The Gate NU, implemented as a MLP, processes domain information to generate attention weights. Given input feature embeddings $x_{\mathrm{feat}} \in \mathbb{R}^{c\times d}$ and domain embedding $x_{\mathrm{dom}}\in\mathbb{R}^d$, where $c$ represents feature columns and $d$ denotes embedding dimension, the Gate NU $\mathrm{gate}(\cdot)$ performs the following computations:
\begin{align*}
    a &= \mathrm{gate}(x_{\mathrm{dom}}) \in \mathbb{R}^d \\
    z &= \mathrm{flatten}(a \otimes x_{\mathrm{feat}}) \in \mathbb{R}^{cd}.
\end{align*}
The operation first generates domain-specific attention weights $a$ through the Gate NU. These weights modulate feature importance by element-wise multiplication with $x_{\mathrm{feat}}$. The resulting domain-weighted features are then flattened into a single vector $z \in \mathbb{R}^{cd}$, yielding the output representation.

This domain-aware feature transformation helps the model adapt to domain-specific characteristics while maintaining the underlying feature semantics. The learned attention weights allow the model to selectively emphasize or suppress features based on their relevance to each domain, improving cross-domain generalization.

Partitioned Norm (PN) extends traditional Batch Normalization (BN) to address domain-specific challenges in feature normalization. Standard BN normalizes features using batch statistics:
\begin{align*}
    z_{\mathrm{out}} = \gamma\frac{z_{\mathrm{in}} - \mu}{\sqrt{\sigma^2+\epsilon}} + \beta,
\end{align*}
where $\mu$ and $\sigma$ are computed from moving averages across training batches. However, when data from multiple domains is mixed, these statistics become unreliable due to domain distribution shifts. This leads to suboptimal normalization and potentially degraded model performance.

To overcome this limitation, PN introduces domain-specific normalization parameters $\{\gamma_k, \beta_k\}$ and statistics $\{\mu_k, \sigma_k\}$. The domain-specific statistics are collected only within their respective domains, ensuring more accurate normalization. The modified normalization equation becomes:
\begin{align*}
    z_{\mathrm{out}} = \gamma\cdot\gamma_k\frac{z_{\mathrm{in}} - \mu_k}{\sqrt{\sigma_k^2+\epsilon}} + \beta + \beta_k.
\end{align*}
This combines global normalization parameters ($\gamma, \beta$) with domain-specific adjustments ($\gamma_k, \beta_k$). Domain-specific scaling $\gamma_k$ and shifting $\beta_k$ enable optimal domain transformations. Using domain-specific statistics $\mu_k$ and $\sigma_k$, PN accurately captures each domain's statistics, enhancing feature representation and model performance.

\paragraph{\textbf{TIN Module}}
Our model uses the Temporal Interest Network (TIN) \cite{zhou2023temporal} to encode temporal and semantic relationships in user behaviors. TIN employs target-aware temporal encoding, attention, and representation to capture correlations between past behaviors and target items. Building on prior work \cite{zhou2023temporal}, we enhance TIN with user ID embeddings to improve modeling of behavior sequences with target game features, thereby enhancing the capture of user-specific patterns while retaining TIN's core strengths.

\paragraph{\textbf{Tower Layer}}
The tower layer integrates and processes various embeddings to generate final predictions. It employs AdaSparse \cite{yang2022adasparse} for domain-specific neuron filtering, adapting the network structure for each domain. To handle the long-tailed distribution of user values, we adopt a zero-inflated lognormal distribution \cite{wang2019deep} modeling approach. We refer our readers to Appendix \ref{appendix:ziln} for detailed explanations about ZILN.

\subsection{Pareto Optimization}
\label{sec:pareto_optim}
Training a model that excels across all time horizons in multi-period prediction is challenging.
To address this challenge, we train the aforementioned components with Pareto optimization techniques. Our applied Pareto optimization method consists of two primary stages: (1) Non-Dominating Gradient Descent, and (2) Optimal Search. In the first stage, we identify a non-dominating gradient, which balances the direction of gradients from multiple tasks (\textit{i.e.,} 3/7/30-Value). It helps alleviate the gradient conflict problem. Then, we are able to obtain a Pareto optimal model via descending along the non-dominating gradient. In the second stage, we search along the possible Pareto front and find a model that performs well and is balanced across all time horizons.

In Pareto Optimization, the objective function set is defined as $\mathcal{L}_{\mathrm{LTV}} = \{l_t\}_{t\in \mathcal{T}}$, where $l_t$ is the objective function of the $t$-Value task. In math, $l_t$ can be described as
\begin{align*}
    l_t= \frac{1}{\|\mathbf{B}\|}\sum_{(p,q)} (y_{t-\mathrm{value}}^{u_p,i_q} - f(u_p,x_{u_p},b_{u_p},i_q,x_{i_q} | \Theta_{t-\mathrm{value}}))^2,
\end{align*}
where $\mathbf{B}$ is the batch of data samples and $(p,q)\in\mathbf{B}$.

\textbf{Non-Dominating Gradient Descent.} Let $g_i=\nabla l_i$ denote the gradient of the $i$-th objective function. Consequently, we define $G= \nabla \mathcal{L} = [g_1,\ldots, g_m]$ by back-propagating the derivatives from the respective objectives. To navigate toward the Pareto front, \citet{mgda} established that the descent direction $d$ can be located within the convex hull of the gradients, represented as $d \in \mathcal{CH}_{\boldsymbol{x}}:= \{ G \boldsymbol{\beta} \}$, where $\boldsymbol{\beta} \in \mathcal{S}^m$ is a vector in the $m$-dimensional simplex. To determine the \emph{Non-Dominating Descent Direction} $d_{nd} = G\boldsymbol{\beta}^*$, we are motivated by Pareto optimization~\citep{mahapatra2020multi} to solve the following \emph{Quadratic Programming} (QP) problem:
\begin{equation}
\begin{aligned}
 \boldsymbol{\beta}^* = & \underset{\|\boldsymbol{\beta}\|_1 \leqslant 1}{\arg \min }\left\|G^\top G \boldsymbol{\beta}-\mathbf{a}\right\|^2 \\
& \text { s.t. } \boldsymbol{\beta}^\top G^\top g_j \geqslant 0 \quad \forall j \in \mathrm{J}=\left\{\begin{array}{cc}
\mathrm{J}^* & \mathrm{KL}\left(\mathcal{L}_{\mathrm{LTV}} \odot \boldsymbol{\lambda} | \boldsymbol{1} \right) \leqslant \epsilon \\
{[m]} & \mathrm{KL}\left(\mathcal{L}_{\mathrm{LTV}} \odot \boldsymbol{\lambda} | \boldsymbol{1} \right) > \epsilon
\end{array},\right. \\
& \text { where } \quad \mathrm{J}^*=\left\{j \in[m] \mid j=\arg \max _{j^{\prime} \in[m]} l_{j^{\prime}} \lambda_{j^{\prime}}\right\},
\end{aligned}
\label{eq:qp}
\end{equation}
where $\mathbf{a}$ is the anchoring direction~\citep{mahapatra2020multi}, $\epsilon>0$ is a hyperparameter, and $\boldsymbol{\lambda}\in \mathbb{R}^m$ is a predefined weight vector that indicates the importance of each period. Then, we calculate the non-dominating direction $d_{nd}=G\boldsymbol{\beta}^*$ and update the model. Therefore, we can yield a Pareto optimal model conditioned on the given importance vector.

\textbf{Optimal Search.} In practice, we expect the accuracy of each period to be balanced. In other words, the weight vector is distributed around the unit vector.  Therefore, we define the weight vector $\boldsymbol{\lambda}$ on the unit sphere with inclination angle and azimuth angle ranging from $\frac{\pi}{6}$ to $\frac{\pi}{3}$. We first 
randomly generate two variables $u$ and $v$ ranging from $\frac{1}{3}$ to $\frac{2}{3}$, then compute the inclination angle $\theta = \frac{\pi}{2} u$ and azimuth angle $\phi = \arccos{v}$. Finally, we have the weight vector $\boldsymbol{\lambda} = [\lambda_1,\lambda_2,\lambda_3]$ as follows
\begin{equation*}
\left\{
     \begin{array}{lr}
     \lambda_1=\sin{\phi}\cos {\theta}, &  \\
     \lambda_2=\sin{\phi}\sin {\theta}, \\
     \lambda_3=\cos {\phi}. &  
     \end{array}
\right.
\end{equation*}
The Pareto optimization for multi-horizon LTV is illustrated in Algorithm~\ref{alg:pareto-optimization}. The computation of $G^T G$ has a time complexity of $O(m^2 n)$, where $n$ is the dimension  of the gradients. Utilizing the most efficient QP solver available~\citep{zhang2021wide}, the runtime is $O(m^3)$. Given that in deep networks, it is typically the case that  $n \gg  m$, the implementation of Pareto optimization does not notably raise the computational expense associated with calculating the non-dominating gradient. Given that the \ourmeth~components are lightweight, the whole framework remains efficient.

%% file: tex/experiment.tex
\section{Experiments}

\input{table/main_results}

\subsection{Dataset: Wechat Mini-Game Recommendation}
Given the absence of public datasets for LTV prediction in a general game platform, we create an industry dataset collected from the WeChat mini games platform to perform offline evaluation. The dataset contains information at three levels: user, game, and behavior. At the user level, we considered each user's age, gender, city of residence, total number of payments, and other factors. At the game level, we included factors such as game category (\textit{e.g.,} casual), battle type (\textit{e.g.,} level-based), market type (\textit{e.g.,} simulation), and game theme (\textit{e.g.,} music). At the behavior level, we consider the user's purchase history for games. To conduct LTV prediction, we retrieve all users who made payments for WeChat mini-games due to advertising over a six-month period and collect their cumulative payments for 3-day, 7-day, and 30-day intervals. Ultimately, we obtain 3,730,392 LTV samples, representing newly registered users for a specific game. On average, there are approximately 21,000 new registered users each day. For offline evaluation purposes, we partitioned the WeChat mini game dataset into training, validation, and test sets with a distribution ratio of 7:2:1.

\subsection{Evaluation Protocols}
We focus on assessing the performance of a model designed to predict user LTV by distinguishing between high-value and low-value users. To this end, we apply three key evaluation metrics: the Normalized Mean Average Error (NMAE), the Area Under the Curve (AUC) and the Normalized Gini Coefficient (N-GINI). AUC evaluates the classification performance of a model. A higher AUC value indicates that the model is more effective at distinguishing between consumers and non-consumers. However, the AUC does not reflect the accuracy of user rankings based on predicted LTV. Therefore, following \cite{zhang2023out,zhou2024cross}, we further introduce N-GINI, a metric that measures the model's ability to rank users accurately according to their predicted LTV. The N-GINI metric ranges from 0 to 1, with higher values indicating greater consistency between the rankings based on predicted LTV and those based on real LTV.

\subsection{Baselines}
Since LTV is a task based on time series behavior, we not only consider methods specifically designed for LTV but also introduce advanced time series forecasting methods.

Time Series Forecasting (TSF) category:

\ding{172} TCN \cite{bai2018empirical} is developed for sequence modeling with a convolutional neural network (CNN) backbone.

\ding{173} LSTM \cite{10.1162/neco.1997.9.8.1735} is a classical and effective sequence modeling method based on recurrent neural networks.

\ding{174} Informer \cite{zhou2021informer} is a variant of the conventional Transformer that introduces probSparse self-attention to significantly reduce computational complexity.

\ding{175} ARIMA \cite{box2015time} is a renowned method for time series forecasting that combines autoregressive and moving average components with differencing to achieve stationarity.

Life-time Value (LTV) Prediction category:

\ding{172} GateNet \cite{huang2020gatenet} employs a gating module to select relevant latent information from the features, enhancing LTV prediction.

\ding{173} TSUR \cite{xing2021learning}, a temporal-structural user representation model, enhances both temporal and structural encoding for LTV prediction.

\ding{174} CDLtvS \cite{zhou2024a} is a novel cross domain method, which applies rich cross domain information to enhance user representations for LTV prediction.

\ding{175} ADSNet \cite{wang2024adsnet} is a deep neural network based on ordinal classification for LTV prediction. It comprises an encoding layer, an expert layer, and a tower layer.

\ding{176} ZILN \cite{wang2019deep} is a zero-inflated lognormal loss function to address the imbalanced problem for LTV prediction.

\ding{177} DeepFM \cite{guo2017deepfm} combines the factorization machine model and deep neural networks to learn both low- and high-order feature interactions. 

\ding{178} Kuaishou \cite{li2022billion} introduces a module that models the ordered dependencies between LTVs of different time spans. Also, a multi-expert strategy is used to transform the severely imbalanced distribution modeling problem into a series of relatively balanced sub-distribution modeling problems.

\subsection{Overall Performance}
Predicting user life-time value is important for ad bidding in mini-game advertising, which helps advertisers set better bids for ad spaces, increasing their return on investment. Moreover, different time windows reveal short-term and long-term user value. Short-term (3-day) shows immediate responses, while long-term (30-day) shows ongoing engagement. This helps Tencent adjust the advertising strategies. Therefore, to ensure a fair comparison, we guarantee that all methods predict the LTV results for the three time windows using a single backbone.

In Table 1, we summarize the performance of all methods in predicting LTV, measured by NMAE, AUC, and GINI. Based on these results, we have drawn the following conclusions:
\paragraph{\textbf{Effectiveness}} Our method serves as a strong baseline, consistently outperforming all other baselines by a large margin across the three metrics. Specifically, compared to the second best methods, relative improvements of our method across the three metrics are 14.0\%, 3.6\% and 1.6\%. Overall, models specifically designed for the LTV prediction task tend to outperform classic time series forecasting models. This could be attributed to the prior emphasis on resolving fundamental LTV-specific challenges, including imbalance, domain shift, and sparsity.
\paragraph{\textbf{Ability to balance multiple objectives}}
The results show that most methods are unable to accurately predict the LTV for all three time windows simultaneously. For example, in terms of AUC, GateNet performs well in the 3-day prediction (ranking second) but struggles with the 7-day (ranking eighth) and 30-day (ranking sixth) LTV prediction. It is worth noting that in our method, although the three prediction objectives share a common backbone, optimal results can be attained for each of them.

\begin{figure*}[htbp] 
\centering
\includegraphics[width=\textwidth]{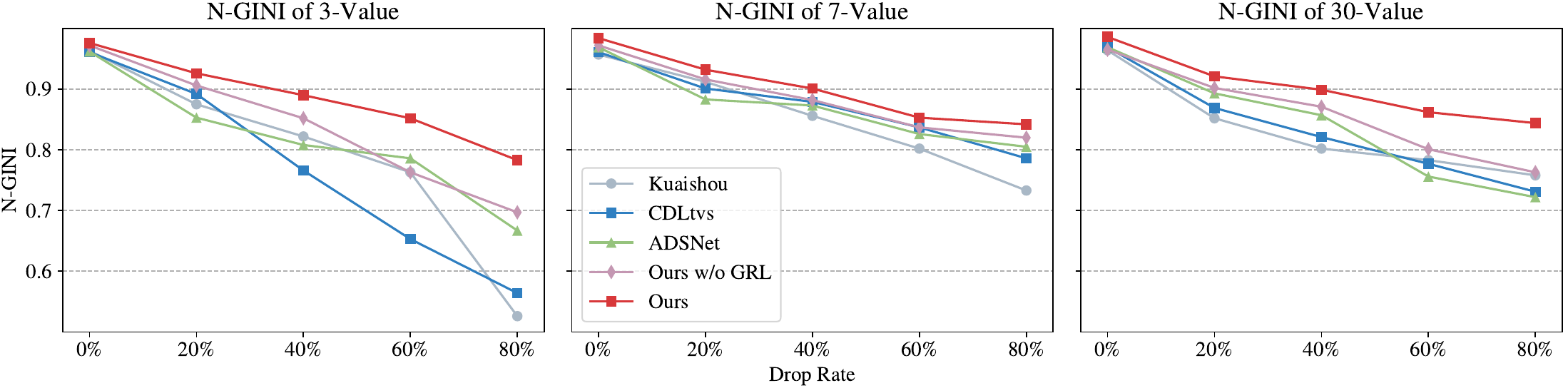}
\caption{Effectiveness of the graph representation learning. We present
the changes in N-GINI values of different methods when dropping a certain ratio of training data.}
\label{fig:grp-effective}
\end{figure*}

\subsection{Effectiveness of Graph Representation Learning}
In this paper, we perform Graph Representation Learning (GRL) on a bipartite graph of historical payments, which aims to alleviate the data sparsity issue of LTV. Here, we verify that the performance of our method is least sensitive to data sparsity. In Figure~\ref{fig:grp-effective}, we present the changes in N-GINI values of different methods when we drop a certain ratio of training data. It is evident that the performance of our method declines the slowest, and as the drop ratio increases, the gap between our method and the second-best method widens significantly. Notably, without the GRL configuration, our method (often) still ranks second, but its ability to combat data sparsity significantly declines.

\subsection{Effectiveness of Pareto Optimization}
\begin{figure}[htb]
    \centering
    \includegraphics[width=0.98\linewidth]{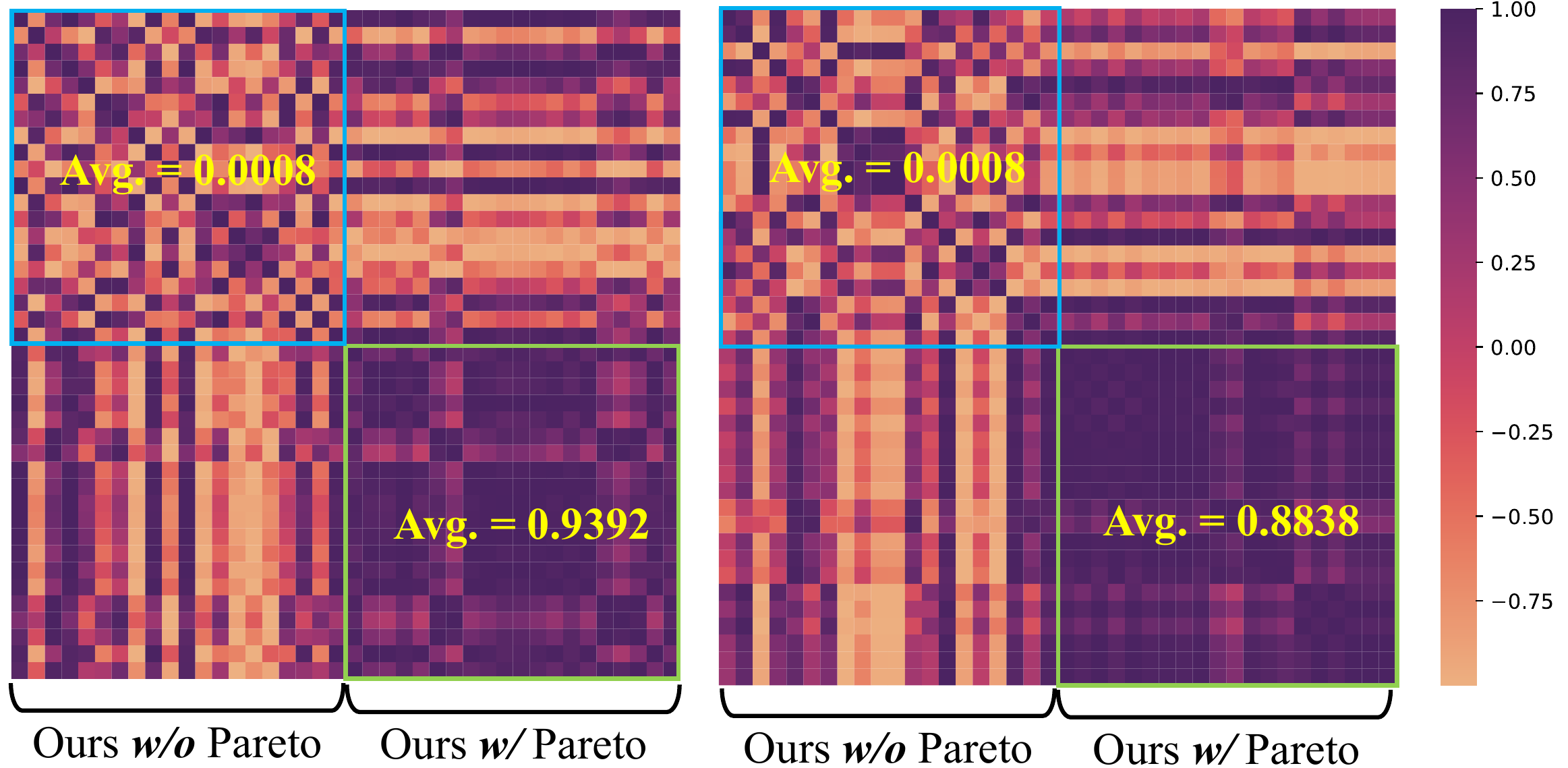}
    \caption{Effectiveness of the Pareto optimization. The left and right figures represent the cases of using and not using the graph representation learning strategy, respectively.}
    \label{fig:pareto-effective}
\end{figure}

To assess the impact of the Pareto optimization strategy implemented in our method, we perform a in-depth analysis based on the volatility of the results. A commonly acknowledged preliminary finding is that when a model has difficulty effectively balancing multiple optimization objectives, altering the training random seed can result in substantial fluctuations in the performance. For evaluation, we perform 20 training runs with different random initializations for both the settings w/ and w/o Pareto, and calculate the testing AUC values for 3-, 7-, and 30- day. As a result, we obtain 40 three-dimensional AUC vectors. Finally, Figure~\ref{fig:pareto-effective} shows the correlation matrix of the 40 vectors. It is evident that the lower right corner of the heatmap is noticeably darker, indicating a higher level of correlation. In contrast, if Pareto is not used, multiple training runs will lead to different results with weak correlations.

\subsection{Online A/B Test}
We validate the effectiveness of our model in the WeChat mini-game recommendation scenario based on two key industrial concerns: accuracy and stability.
\paragraph{\textbf{Accuracy}}
From an industrial perspective, the evaluation metrics for accuracy are the Gross Merchandise Value (GMV) and Lifetime Value (LTV) indicators in online experiments. We denote the 3-Value as $\text{LTV}_3$ and define the corresponding GMV as $\text{GMV}_3 = \text{LTV}_3 / \text{ROI}_3$, where the $\text{ROI}_3$ is determined by the advertiser and is almost fixed based on the desired profit margin. For risk mitigation and speed, we perform UV sampling. We create three control groups and one experimental group, with each group conducting a 5\% traffic sampling. The control groups are: the original baseline framework, GRePO-LTV without Pareto, and GRePO-LTV without GRL. In Table~\ref{table:abtest}, we present the relative improvements or reductions of all groups compared to the original baseline, with respect to the LTV and GMV metrics. Results show that our method achieves consistent improvement across all three kinds of time windows. Additionally, our core designs, Pareto and GRL, have significantly improved performance.
In online industrial scenarios, predicting short-term LTV is often more challenging. It is evident that GRePO-LTV shows the greatest improvement in the short-term (3-day) scenario, which is largely attributed to the Pareto strategy.

\input{table/abtest}

\begin{figure}[htbp] 
\centering
\includegraphics[width=\linewidth]{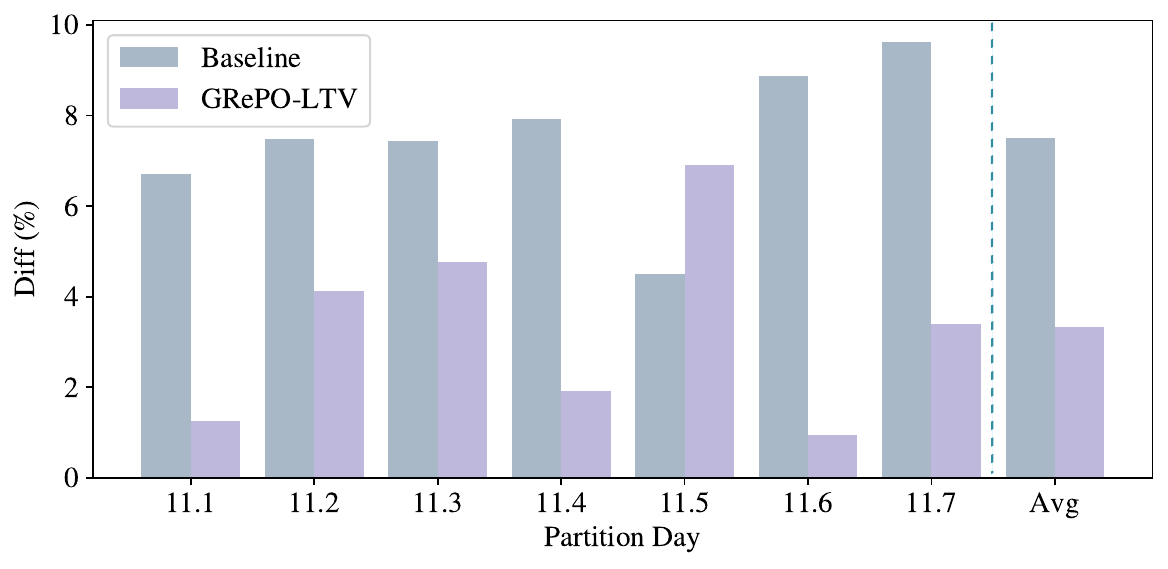}
\caption{Stability analysis comparing \ourmeth~ and baseline over 7 days.}
\label{fig:stability}
\end{figure}

\paragraph{\textbf{Stability}}
Due to the high timeliness of online models, stability poses a significant challenge for LTV in practice. In simple terms, we hope that the next model update does not cause significant fluctuations in the LTV predictions for the fixed samples. The stability evaluation method we employed is as follows: We use two models pushed on adjacent days to make predictions on a batch of fixed samples (same ads, same users, same ad placements), and calculate the differences in predictions. We utilize the following equation as our stability metric, where pLTV$_{1,i}$ represents the prediction result for sample $i$ from the previous day. The larger the Diff, the worse the stability.
\begin{equation*}
    \mathrm{Diff} = \frac{|\sum_i\mathrm{pLTV}_{1,i}-\sum_i\mathrm{pLTV}_{2,i}|}{\sum_i\mathrm{pLTV}_{1,i}}
\end{equation*}
We show the stability results in Figure \ref{fig:stability}. Based on the same sample reflux rate, overall, the Diff value of \ourmeth~ is about half lower than that of the baseline. This demonstrates the superior stability of our method and also indicates the reliability of the accuracy comparison in the online experiment.

%% file: table/main_results.tex
\begin{table*}[htb]
\setlength{\tabcolsep}{2mm}
\centering 
\caption{Performance comparison of state-of-the-art LTV prediction methods on three tasks of different time horizons. The best results are marked in bold. $\pm$ indicates the standard deviation of the three tasks for each metric.}
\label{table:overall}
\resizebox{\textwidth}{!}{
\begin{tabular}{cc|ccc|ccc|ccc|ccc}
\toprule
 \multicolumn{2}{c}{\multirow{2}{*}{\textbf{Method}}}  & \multicolumn{3}{c|}{\textbf{3-Value}}  & \multicolumn{3}{c|}{\textbf{7-Value}}  & \multicolumn{3}{c|}{\textbf{30-Value}}  & \multicolumn{3}{c}{\textbf{Average}} \\
& & NMAE & AUC & N-GINI & NMAE & AUC & N-GINI & NMAE & AUC & N-GINI & NMAE & AUC & N-GINI\\
\midrule
 \multirow{4}{*}{\rotatebox{90}{\textbf{TSF}}} & \textbf{TCN} & 0.355 & 0.675 & 0.932 & 0.318 & 0.702 & 0.946 & 0.327 & 0.709 & 0.941 & $0.333_{\pm0.019}$ & $0.695_{\pm0.015}$ & $0.940_{\pm0.006}$ \\
 & \textbf{LSTM}  & 0.322 & 0.711 & 0.940 & 0.357 & 0.693 & 0.932 & 0.277 & 0.715 & 0.948 & $0.319_{\pm0.040}$ & $0.706_{\pm0.010}$ & $0.940_{\pm0.007}$ \\
 & \textbf{Informer}  & 0.298 & 0.712 & 0.954 & 0.302 & 0.722 & 0.956 & 0.376 & 0.657 & 0.922 & $0.325_{\pm0.044}$ & $0.697_{\pm0.029}$ & $0.944_{\pm0.016}$ \\
 & \textbf{ARIMA}  & 0.339 & 0.702 & 0.937 & 0.319 & 0.712 & 0.941 & 0.298 & 0.728 & 0.953 & $0.319_{\pm0.020}$ & $0.714_{\pm0.011}$ & $0.944_{\pm0.007}$ \\

\midrule
 \multirow{7}{*}{\rotatebox{90}{\textbf{LTV}}} & \textbf{GateNet} & 0.254 & 0.732 & 0.958 & 0.354 & 0.709 & 0.930 & 0.303 & 0.719 & 0.956 & $0.304_{\pm0.050}$ & $0.720_{\pm0.009}$ & $0.948_{\pm0.013}$ \\
 & \textbf{TSUR}  & 0.289 & 0.711 & 0.951 & 0.319 & 0.702 & 0.939 & 0.358 & 0.678 & 0.930 & $0.322_{\pm0.035}$ & $0.697_{\pm0.014}$ & $0.940_{\pm0.009}$  \\
 & \textbf{CDLtvS}  & 0.245 & 0.728 & 0.961 & 0.231 & 0.732 & 0.961 & 0.201 & 0.737 & 0.969 & $0.226_{\pm0.022}$ & $0.732_{\pm0.004}$ & $0.964_{\pm0.004}$ \\
 & \textbf{ADSNet}  & 0.241 & 0.726 & 0.962 & 0.218 & 0.742 & 0.969 & 0.197 & 0.742 & 0.969 & $0.219_{\pm0.022}$ & $0.737_{\pm0.008}$ & $0.967_{\pm0.003}$  \\
 & \textbf{ZILN}  & 0.268 & 0.726 & 0.958 & 0.337 & 0.698 & 0.935 & 0.319 & 0.692 & 0.943 & $0.308_{\pm0.036}$ & $0.705_{\pm0.015}$ & $0.945_{\pm0.010}$  \\
  & \textbf{Kuaishou}  &  0.232 & 0.731 & 0.963 & 0.298 & 0.716 & 0.957 & 0.259 & 0.729 & 0.964 & $0.263_{\pm0.033}$ & $0.725_{\pm0.007}$ & $0.961_{\pm0.003}$  \\
& \textbf{DeepFM}  & 0.270 & 0.727 & 0.960 & 0.258 & 0.736 & 0.967 & 0.320 & 0.706 & 0.945 & $0.283_{\pm0.033}$ & $0.723_{\pm0.013}$ & $0.957_{\pm0.009}$  \\
\midrule
 \multirow{3}{*}{\rotatebox{90}{\textbf{Ours}}} & \textbf{w/o Pareto} & 0.211 & 0.746 & 0.972 & 0.254 & 0.736 & 0.965 & 0.198 & 0.752 & 0.963 & $0.221_{\pm0.029}$ & $0.745_{\pm0.007}$ & $0.967_{\pm0.004}$  \\
 & \textbf{w/o GRL}  & 0.214 & 0.744 & 0.972 & 0.209 & 0.753 & 0.972 & 0.201 & 0.757 & 0.965 & $0.208_{\pm0.007}$ & $0.751_{\pm0.005}$ & $0.970_{\pm0.003}$ \\
 & \textbf{\ourmeth (Ours)}  & \textbf{0.193} & \textbf{0.759} & \textbf{0.976} & \textbf{0.183} & \textbf{0.763} & \textbf{0.984} & \textbf{0.188} & \textbf{0.768} & \textbf{0.986} & $\textbf{0.188}_{\pm0.005}$ & $\textbf{0.763}_{\pm0.004}$ & $\textbf{0.982}_{\pm0.004}$ \\

\bottomrule
\end{tabular}
}
\end{table*}

%% file: table/abtest.tex
\begin{table*}[htb]
\centering
\caption{A/B test results based on the WeChat mini-game traffic. We report the relative changes; darker blue indicates a greater improvement, while darker red indicates a greater decrease.}
\label{table:abtest}
\resizebox{0.75\textwidth}{!}{%
\begin{tabular}{l|cccccccc}
\toprule
\textbf{Groups} &
  LTV$_3$ &
  \multicolumn{1}{c|}{GMV$_3$} &
  LTV$_7$ &
  \multicolumn{1}{c|}{GMV$_7$} &
  LTV$_{30}$ &
  \multicolumn{1}{c|}{GMV$_{30}$} &
  \multicolumn{1}{c}{LTV$_\mathrm{avg}$} &
  \multicolumn{1}{c}{GMV$_\mathrm{avg}$} \\ \midrule
Original Baseline &
  \multicolumn{8}{c}{--- (Reference)} \\ \midrule
Ours w/o GRL &
  \cellcolor{blue!25}+2.15\% &
  \multicolumn{1}{c|}{\cellcolor{blue!29.5}+2.47\%} &
  \cellcolor{blue!40}+3.77\% &
  \multicolumn{1}{c|}{\cellcolor{blue!34}+3.27\%} &
  \cellcolor{blue!18}+1.51\% &
  \multicolumn{1}{c|}{\cellcolor{blue!18}+1.51\%} &
  \cellcolor{blue!30}2.4\% &
  \cellcolor{blue!29}2.4\% \\
Ours w/o Pareto &
  \cellcolor{red!10}-0.37\% &
  \multicolumn{1}{c|}{\cellcolor{red!15}-0.72\%} &
  \cellcolor{blue!13}+1.05\% &
  \multicolumn{1}{c|}{\cellcolor{blue!15}+1.25\%} &
  \cellcolor{blue!4}+0.17\% &
  \multicolumn{1}{c|}{\cellcolor{red!8}-0.19\%} &
  \cellcolor{blue!8}0.28\% &
  \cellcolor{blue!4}0.11\% \\
GRePO-LTV (ours) &
  \cellcolor{blue!55}+9.91\% &
  \multicolumn{1}{c|}{\cellcolor{blue!53}+9.83\%} &
  \cellcolor{blue!47}+7.80\% &
  \multicolumn{1}{c|}{\cellcolor{blue!47.6}+7.93\%} &
  \cellcolor{blue!46}+7.73\% &
  \multicolumn{1}{c|}{\cellcolor{blue!45.6}+7.60\%} &
  \cellcolor{blue!50.9}8.4\% &
  \cellcolor{blue!50.72}8.4\% \\ \bottomrule
\end{tabular}%
}
\end{table*}

%% file: tex/lessons.tex
\section{Lessons Learned from Deployment}

We have two takeaways during the deployment of GRePO-LTV.

\paragraph{\textbf{Time Span for Graph Data.}}
To effectively address the data paucity problem, the graph used for representation learning should contain a sufficient number of edges to enable successful message passing, as multi-hop information can be deeply hidden within the historical data.
Consequently, we explore the entire lifespan of each game to gather adequate graph data.
This approach ensures the highest quality of user and game embeddings.

\paragraph{\textbf{Accounting for Outliers on Special Days}}
Certain events in a game's lifecycle, such as the launch day or limited-time events, can significantly impact player behavior and generate outliers in the data distribution. During these periods, the existing trained model may not accurately represent or predict the current state of the game. To mitigate this issue and ensure the model adapts to these rapid changes, it is recommended to increase the frequency of data collection and model retraining when such events occur. By updating the model more frequently during these times, it can better capture and adjust to the sudden shifts in player behavior and game dynamics, ultimately improving its predictive accuracy and robustness in the face of outlier events.

%% file: tex/related_work.tex
\section{Related Work}

\paragraph{\textbf{Lifetime Value Prediction}}
Forecasting the potential revenue of users is critical for online marketing, which directly impacts the effectiveness of advertisements. Early methods rely on simple metrics (e.g., average purchase frequency) and statistical models such as RFM~\cite{fader2005rfm} and Pareto/NBD models~\cite{fader2005counting,bemmaor2012modeling}, which focus on historical data and are insufficient to capture complex, non-linear relationships in customer behavior. Machine learning models have been a promising direction to address the above weaknesses and multiple studies have explored them for LTV prediction, including random forest~\cite{vanderveld2016engagement}, XGBoost~\cite{drachen2018or}, CNNs~\cite{chen2018customer}, and RNNs~\cite{xing2021learning,bauer2021improved}. However, the extreme data sparsity problem has prevented them from achieving better performance, which has gained considerable attention in recent studies. They can be broadly categorized into two classes: (1) leveraging the LTV data from various domains and employing cross-domain learnings to prevent negative transfer~\cite{zhou2024cross,wang2024adsnet,weng2024optdist,su2023cross,li2022billion}. For example, CDLtvS~\cite{zhou2024cross} utilizes the LTV data from E-commerce and logistics service domains as well as designs a various-level alignment mechanism to ensure knowledge transfer; (2) Leverage the rich side information to enhance the quality of user representations~\cite{10.1145/3580305.3599871,xing2021learning,yang2023feature,zhang2023out,liu2020modelling}. MDLUR~\cite{10.1145/3580305.3599871} employs the additional user portrait and behavior data from multiple sources to enrich the user representations. Moreover, ExpLTV~\cite{zhang2023out} incorporates labels on whether users are highly spent on in-game purchases.
In this work, through graph representation learning techniques, we creatively incorporate collaborative signals to obtain better user and game representations.

\paragraph{\textbf{Pareto Optimization}}
Pareto optimization has emerged as a powerful tool for modeling trade-offs among multiple objectives in various machine learning tasks, such as neural architecture search~\cite{lomurno2021pareto} and long-tailed recognition~\cite{DBLP:conf/iclr/Zhou0Z024}. In recent years, several studies~\cite{jin2024pareto,lin2019pareto,wu2022multi,xie2021personalized} have explored its application to multi-objective recommendation systems. PE-LTR~\cite{lin2019pareto} pioneered the introduction of a Pareto-efficient framework with theoretical guarantees for learning multiple conflicting objectives. Similarly, PAPERec~\cite{xie2021personalized} proposed a personalized Pareto-efficient framework that jointly predicts the dwell time and click-through rate (CTR)~\cite{min2023scenario} metrics for users. Building upon these advancements, this work leverages Pareto optimization techniques to mitigate the challenges associated with predicting values across different time horizons, thereby enhancing the effectiveness of the recommendation system.
 

%% file: tex/conclusion.tex
\section{Conclusion}

We introduced \ourmeth, a novel framework for predicting user lifetime value in mini-game advertising. By employing graph representation learning to capture collaborative signals and Pareto optimization to balance short- and long-term accuracy, \ourmeth~overcomes key challenges like data sparsity and multi-goal conflicts. Our method shows significant improvements in both offline experiments and online A/B testing, highlighting its potential for optimizing game advertising strategies.

%% file: tex/acknowledgment.tex
\begin{acks}
This research was supported by Tencent Inc. under project number AMS RBFR2024002. We extend our sincere gratitude to Tencent Inc. for this support.
\end{acks}

%% file: tex/appendix.tex
\setcounter{section}{0}
\setcounter{figure}{0}
\setcounter{table}{0}
\setcounter{algorithm}{0}

\renewcommand{\thefigure}{A\arabic{figure}}
\renewcommand{\theHfigure}{A\arabic{figure}}
\renewcommand{\thetable}{A\arabic{table}}
\renewcommand{\theHtable}{A\arabic{table}}
\renewcommand{\thealgorithm}{A\arabic{algorithm}}
\renewcommand{\theHalgorithm}{A\arabic{algorithm}}

\section{Pareto Optimization for Multi-horizon LTV}

\begin{algorithm}[htb]
  \caption{Pareto optimization for multi-horizon LTV}
  \label{alg:pareto-optimization}
  \begin{algorithmic}[1] 
    \Require Step size $\eta>0$.
    \For{$k=1,\ldots,K$}
      \State Randomly generate $u$ and $v$ ranging from $\frac{1}{3}$ to $\frac{2}{3}$.
      \State Compute angle $\theta = \frac{\pi}{2} u$, $\phi = \arccos{v}$.
      \State Compute weight $\boldsymbol{\lambda}=[\sin{\phi}\cos {\theta}, \sin{\phi}\sin {\theta}, \cos {\phi}]$.
      \For{$t=1,\ldots,T$}
        \State Initialization.
        \State Compute gradients of objectives: $G= \nabla \mathcal{L}$.
        \State Determine $\boldsymbol{\beta}^*$ by solving QP problem in Eq. \ref{eq:qp}.
        \State Calculate non-dominating gradient $d_{nd} = G\boldsymbol{\beta}^*$.
        \State Update the model with $\eta d_{nd}$.
      \EndFor
      \State Output model $F^k$.
    \EndFor
    \Ensure $F^*$ from $[F^1, \cdots, F^K]$.
  \end{algorithmic}
\end{algorithm}

\section{Zero-Inflated Lognormal Distribution}
\label{appendix:ziln}
\subsection{Overview}
The Zero-Inflated Lognormal (ZILN) distribution is a statistical model designed for datasets that exhibit a high frequency of zero values alongside positive, continuous data that are typically right-skewed \cite{wang2019deep}f. Standard continuous probability distributions, including the conventional lognormal distribution, are often inadequate for such data as they cannot inherently model a significant point mass at zero \cite{crow1987lognormal}. The ZILN model addresses this by conceptualizing the data as arising from a mixture of two distinct processes: one that generates zero values and another that generates positive values following a lognormal distribution. 

\subsection{The Lognormal Component}
The foundation for the positive, continuous part of the ZILN distribution is the standard lognormal distribution. A random variable $X$ is said to follow a lognormal distribution if its natural logarithm, $Y = \ln(X)$, is normally distributed with mean $\mu$ and variance $\sigma^2$. The probability density function (PDF) of a 2-parameter lognormal distribution is given by \cite{crow1987lognormal}:
$$ f_{LN}(x | \mu, \sigma^2) = \frac{1}{x\sigma\sqrt{2\pi}} \exp\left(-\frac{(\ln x - \mu)^2}{2\sigma^2}\right), \quad \text{for } x > 0 $$
The lognormal distribution is defined exclusively for positive real values ($X > 0$) and is characterized by its inherent right-skewness. Crucially, it cannot assign any probability mass to $x=0$.

\subsection{Structure of the ZILN Distribution}
The ZILN distribution is a specialized mixture model that combines two processes:
\ding{172} A binary process, typically modeled as a Bernoulli trial, determines whether an observation is a `structural' or `excess' zero. This occurs with a certain probability, denoted as $\pi$. 
\ding{173} If the observation is not an `excess' zero (which happens with probability $1-\pi$), its value is drawn from a lognormal distribution with parameters $\mu$ (mean of the log-transformed positive values) and $\sigma^2$ (variance of the log-transformed positive values). 

Since the lognormal component is strictly positive, any observed zero value in a ZILN-distributed dataset is attributed to the zero-generating (inflation) part of the model.

\subsection{Mathematical Definition}
Let $Y$ be a random variable following a Zero-Inflated Lognormal distribution. The distribution is characterized by three parameters: the zero-inflation probability $\pi$ ($0 \le \pi < 1$), and the lognormal parameters $\mu$ (mean of $\ln Y$ for $Y>0$) and $\sigma^2$ (variance of $\ln Y$ for $Y>0$, with $\sigma > 0$).
The probability density function (PDF) of the ZILN distribution, $g(y; \pi, \mu, \sigma^2)$, is expressed in a piecewise form as:
$$
g(y; \pi, \mu, \sigma^2) =
\begin{cases}
\pi & \text{if } y = 0 \\
(1-\pi) \frac{1}{y\sigma\sqrt{2\pi}} \exp\left(-\frac{(\ln y - \mu)^2}{2\sigma^2}\right) & \text{if } y > 0
\end{cases}
$$
This formulation shows a discrete probability mass $\pi$ at $y=0$ and a scaled lognormal density for $y>0$.

\section{Performance for Different User Groups}

\input{table/ltv_bias}

We group our users according to their LTVs and report the average prediction bias ($\frac{|\mathrm{LTV}_\mathrm{pred}-\mathrm{LTV}_\mathrm{true}|}{\mathrm{LTV}_\mathrm{true}}$) as a percentage. Results are shown in Table \ref{tab:ltv_bias_according_to_user_distribution}.

The model exhibits a systematic bias pattern, consistently overestimating users with lower LTVs while underestimating those with higher LTVs. Interestingly, this bias is less extreme for long-term LTV predictions compared to short-term ones. The transition from overestimation to underestimation typically occurs when the LTV falls within the 500 to 3000 range.

%% file: table/ltv_bias.tex
\begin{table}[h!] 
\centering 
\resizebox{\columnwidth}{!}{%
\begin{tabular}{|l|c|c|c|}
\hline
\textbf{LTV Range} & \textbf{User Distribution} & \textbf{3-Value Bias} & \textbf{30-Value Bias} \\
\hline
(0.1,6] & 7.99\% & 344.04\% & 188.26\% \\
(6,18] & 11.80\% & 298.85\% & 193.79\% \\
(18,30] & 8.78\% & 532.61\% & 183.10\% \\
(30,100] & 22.20\% & 369.55\% & 144.18\% \\
(100,500] & 20.52\% & 19.30\% & 52.79\% \\
(500,3000] & 16.71\% & -1.44\% & 19.46\% \\
(3000,20000] & 8.73\% & -27.88\% & -14.77\% \\
(20000,$\infty$) & 3.26\% & -49.80\% & -46.50\% \\
\hline
\end{tabular}%
}
\caption{Average LTV prediction bias.} 
\label{tab:ltv_bias_according_to_user_distribution} 
\end{table}

%% file: main.bbl

\begin{thebibliography}{52}


\ifx \showCODEN    \undefined \def \showCODEN     #1{\unskip}     \fi
\ifx \showISBNx    \undefined \def \showISBNx     #1{\unskip}     \fi
\ifx \showISBNxiii \undefined \def \showISBNxiii  #1{\unskip}     \fi
\ifx \showISSN     \undefined \def \showISSN      #1{\unskip}     \fi
\ifx \showLCCN     \undefined \def \showLCCN      #1{\unskip}     \fi
\ifx \shownote     \undefined \def \shownote      #1{#1}          \fi
\ifx \showarticletitle \undefined \def \showarticletitle #1{#1}   \fi
\ifx \showURL      \undefined \def \showURL       {\relax}        \fi
\providecommand\bibfield[2]{#2}
\providecommand\bibinfo[2]{#2}
\providecommand\natexlab[1]{#1}
\providecommand\showeprint[2][]{arXiv:#2}

\bibitem[Bai et~al\mbox{.}(2018)]%
        {bai2018empirical}
\bibfield{author}{\bibinfo{person}{Shaojie Bai}, \bibinfo{person}{J~Zico Kolter}, {and} \bibinfo{person}{Vladlen Koltun}.} \bibinfo{year}{2018}\natexlab{}.
\newblock \showarticletitle{An empirical evaluation of generic convolutional and recurrent networks for sequence modeling}.
\newblock \bibinfo{journal}{\emph{arXiv preprint arXiv:1803.01271}} (\bibinfo{year}{2018}).
\newblock


\bibitem[Bauer and Jannach(2021)]%
        {bauer2021improved}
\bibfield{author}{\bibinfo{person}{Josef Bauer} {and} \bibinfo{person}{Dietmar Jannach}.} \bibinfo{year}{2021}\natexlab{}.
\newblock \showarticletitle{Improved customer lifetime value prediction with sequence-to-sequence learning and feature-based models}.
\newblock \bibinfo{journal}{\emph{TKDD}} \bibinfo{volume}{15}, \bibinfo{number}{5} (\bibinfo{year}{2021}), \bibinfo{pages}{1--37}.
\newblock


\bibitem[Bemmaor and Glady(2012)]%
        {bemmaor2012modeling}
\bibfield{author}{\bibinfo{person}{Albert~C Bemmaor} {and} \bibinfo{person}{Nicolas Glady}.} \bibinfo{year}{2012}\natexlab{}.
\newblock \showarticletitle{Modeling purchasing behavior with sudden “death”: A flexible customer lifetime model}.
\newblock \bibinfo{journal}{\emph{Management Science}} \bibinfo{volume}{58}, \bibinfo{number}{5} (\bibinfo{year}{2012}), \bibinfo{pages}{1012--1021}.
\newblock


\bibitem[Box et~al\mbox{.}(2015)]%
        {box2015time}
\bibfield{author}{\bibinfo{person}{George~EP Box}, \bibinfo{person}{Gwilym~M Jenkins}, \bibinfo{person}{Gregory~C Reinsel}, {and} \bibinfo{person}{Greta~M Ljung}.} \bibinfo{year}{2015}\natexlab{}.
\newblock \bibinfo{booktitle}{\emph{Time series analysis: forecasting and control}}.
\newblock \bibinfo{publisher}{John Wiley \& Sons}.
\newblock


\bibitem[Chang et~al\mbox{.}(2023)]%
        {chang2023pepnet}
\bibfield{author}{\bibinfo{person}{Jianxin Chang}, \bibinfo{person}{Chenbin Zhang}, \bibinfo{person}{Yiqun Hui}, \bibinfo{person}{Dewei Leng}, \bibinfo{person}{Yanan Niu}, \bibinfo{person}{Yang Song}, {and} \bibinfo{person}{Kun Gai}.} \bibinfo{year}{2023}\natexlab{}.
\newblock \showarticletitle{Pepnet: Parameter and embedding personalized network for infusing with personalized prior information}. In \bibinfo{booktitle}{\emph{KDD}}. \bibinfo{pages}{3795--3804}.
\newblock


\bibitem[Chen et~al\mbox{.}(2025)]%
        {chen2025graph}
\bibfield{author}{\bibinfo{person}{Hao Chen}, \bibinfo{person}{Yu Yang}, \bibinfo{person}{Yuanchen Bei}, \bibinfo{person}{Zefan Wang}, \bibinfo{person}{Yue Xu}, {and} \bibinfo{person}{Feiran Huang}.} \bibinfo{year}{2025}\natexlab{}.
\newblock \showarticletitle{Graph Neural Patching for Cold-Start Recommendations}. In \bibinfo{booktitle}{\emph{Australasian Database Conference}}. Springer, \bibinfo{pages}{334--346}.
\newblock


\bibitem[Chen et~al\mbox{.}(2018)]%
        {chen2018customer}
\bibfield{author}{\bibinfo{person}{Pei~Pei Chen}, \bibinfo{person}{Anna Guitart}, \bibinfo{person}{Ana~Fern{\'a}ndez del R{\'\i}o}, {and} \bibinfo{person}{Africa Peri{\'a}nez}.} \bibinfo{year}{2018}\natexlab{}.
\newblock \showarticletitle{Customer lifetime value in video games using deep learning and parametric models}. In \bibinfo{booktitle}{\emph{2018 IEEE international conference on big data (big data)}}. IEEE, \bibinfo{pages}{2134--2140}.
\newblock


\bibitem[Crow and Shimizu(1987)]%
        {crow1987lognormal}
\bibfield{author}{\bibinfo{person}{Edwin~L Crow} {and} \bibinfo{person}{Kunio Shimizu}.} \bibinfo{year}{1987}\natexlab{}.
\newblock \bibinfo{booktitle}{\emph{Lognormal distributions}}.
\newblock \bibinfo{publisher}{Marcel Dekker New York}.
\newblock


\bibitem[Deb et~al\mbox{.}(2016)]%
        {deb2016multi}
\bibfield{author}{\bibinfo{person}{Kalyanmoy Deb}, \bibinfo{person}{Karthik Sindhya}, {and} \bibinfo{person}{Jussi Hakanen}.} \bibinfo{year}{2016}\natexlab{}.
\newblock \showarticletitle{Multi-objective optimization}.
\newblock In \bibinfo{booktitle}{\emph{Decision sciences}}. \bibinfo{publisher}{CRC Press}, \bibinfo{pages}{161--200}.
\newblock


\bibitem[Drachen et~al\mbox{.}(2018)]%
        {drachen2018or}
\bibfield{author}{\bibinfo{person}{Anders Drachen}, \bibinfo{person}{Mari Pastor}, \bibinfo{person}{Aron Liu}, \bibinfo{person}{Dylan~Jack Fontaine}, \bibinfo{person}{Yuan Chang}, \bibinfo{person}{Julian Runge}, \bibinfo{person}{Rafet Sifa}, {and} \bibinfo{person}{Diego Klabjan}.} \bibinfo{year}{2018}\natexlab{}.
\newblock \showarticletitle{To be or not to be... social: Incorporating simple social features in mobile game customer lifetime value predictions}. In \bibinfo{booktitle}{\emph{proceedings of the australasian computer science week multiconference}}. \bibinfo{pages}{1--10}.
\newblock


\bibitem[Désidéri(2012)]%
        {mgda}
\bibfield{author}{\bibinfo{person}{Jean-Antoine Désidéri}.} \bibinfo{year}{2012}\natexlab{}.
\newblock \showarticletitle{Multiple-gradient descent algorithm (MGDA) for multiobjective optimization}.
\newblock \bibinfo{journal}{\emph{Comptes Rendus Mathematique}} \bibinfo{volume}{350}, \bibinfo{number}{5} (\bibinfo{year}{2012}), \bibinfo{pages}{313--318}.
\newblock


\bibitem[Fader et~al\mbox{.}(2005a)]%
        {fader2005rfm}
\bibfield{author}{\bibinfo{person}{Peter~S Fader}, \bibinfo{person}{Bruce~GS Hardie}, {and} \bibinfo{person}{Ka~Lok Lee}.} \bibinfo{year}{2005}\natexlab{a}.
\newblock \showarticletitle{RFM and CLV: Using iso-value curves for customer base analysis}.
\newblock \bibinfo{journal}{\emph{Journal of marketing research}} \bibinfo{volume}{42}, \bibinfo{number}{4} (\bibinfo{year}{2005}), \bibinfo{pages}{415--430}.
\newblock


\bibitem[Fader et~al\mbox{.}(2005b)]%
        {fader2005counting}
\bibfield{author}{\bibinfo{person}{Peter~S Fader}, \bibinfo{person}{Bruce~GS Hardie}, {and} \bibinfo{person}{Ka~Lok Lee}.} \bibinfo{year}{2005}\natexlab{b}.
\newblock \showarticletitle{“Counting your customers” the easy way: An alternative to the Pareto/NBD model}.
\newblock \bibinfo{journal}{\emph{Marketing science}} \bibinfo{volume}{24}, \bibinfo{number}{2} (\bibinfo{year}{2005}), \bibinfo{pages}{275--284}.
\newblock


\bibitem[Guo et~al\mbox{.}(2017)]%
        {guo2017deepfm}
\bibfield{author}{\bibinfo{person}{Huifeng Guo}, \bibinfo{person}{Ruiming Tang}, \bibinfo{person}{Yunming Ye}, \bibinfo{person}{Zhenguo Li}, {and} \bibinfo{person}{Xiuqiang He}.} \bibinfo{year}{2017}\natexlab{}.
\newblock \showarticletitle{DeepFM: a factorization-machine based neural network for CTR prediction}.
\newblock \bibinfo{journal}{\emph{arXiv preprint arXiv:1703.04247}} (\bibinfo{year}{2017}).
\newblock


\bibitem[He et~al\mbox{.}(2020)]%
        {he2020lightgcn}
\bibfield{author}{\bibinfo{person}{Xiangnan He}, \bibinfo{person}{Kuan Deng}, \bibinfo{person}{Xiang Wang}, \bibinfo{person}{Yan Li}, \bibinfo{person}{Yongdong Zhang}, {and} \bibinfo{person}{Meng Wang}.} \bibinfo{year}{2020}\natexlab{}.
\newblock \showarticletitle{Lightgcn: Simplifying and powering graph convolution network for recommendation}. In \bibinfo{booktitle}{\emph{SIGIR}}. \bibinfo{pages}{639--648}.
\newblock


\bibitem[Hochreiter and Schmidhuber(1997)]%
        {10.1162/neco.1997.9.8.1735}
\bibfield{author}{\bibinfo{person}{Sepp Hochreiter} {and} \bibinfo{person}{Jürgen Schmidhuber}.} \bibinfo{year}{1997}\natexlab{}.
\newblock \showarticletitle{Long Short-Term Memory}.
\newblock \bibinfo{journal}{\emph{Neural Computation}} \bibinfo{volume}{9}, \bibinfo{number}{8} (\bibinfo{date}{11} \bibinfo{year}{1997}), \bibinfo{pages}{1735--1780}.
\newblock
\showISSN{0899-7667}
\href{https://doi.org/10.1162/neco.1997.9.8.1735}{doi:\nolinkurl{10.1162/neco.1997.9.8.1735}}
\showeprint{https://direct.mit.edu/neco/article-pdf/9/8/1735/813796/neco.1997.9.8.1735.pdf}


\bibitem[Huang et~al\mbox{.}(2020)]%
        {huang2020gatenet}
\bibfield{author}{\bibinfo{person}{Tongwen Huang}, \bibinfo{person}{Qingyun She}, \bibinfo{person}{Zhiqiang Wang}, {and} \bibinfo{person}{Junlin Zhang}.} \bibinfo{year}{2020}\natexlab{}.
\newblock \showarticletitle{GateNet: gating-enhanced deep network for click-through rate prediction}.
\newblock \bibinfo{journal}{\emph{arXiv preprint arXiv:2007.03519}} (\bibinfo{year}{2020}).
\newblock


\bibitem[Jin et~al\mbox{.}(2024)]%
        {jin2024pareto}
\bibfield{author}{\bibinfo{person}{Jipeng Jin}, \bibinfo{person}{Zhaoxiang Zhang}, \bibinfo{person}{Zhiheng Li}, \bibinfo{person}{Xiaofeng Gao}, \bibinfo{person}{Xiongwen Yang}, \bibinfo{person}{Lei Xiao}, {and} \bibinfo{person}{Jie Jiang}.} \bibinfo{year}{2024}\natexlab{}.
\newblock \showarticletitle{Pareto-based Multi-Objective Recommender System with Forgetting Curve}. In \bibinfo{booktitle}{\emph{CIKM}}. \bibinfo{pages}{4603--4611}.
\newblock


\bibitem[Larivi{\`e}re and Van~den Poel(2005)]%
        {lariviere2005predicting}
\bibfield{author}{\bibinfo{person}{Bart Larivi{\`e}re} {and} \bibinfo{person}{Dirk Van~den Poel}.} \bibinfo{year}{2005}\natexlab{}.
\newblock \showarticletitle{Predicting customer retention and profitability by using random forests and regression forests techniques}.
\newblock \bibinfo{journal}{\emph{Expert systems with applications}} \bibinfo{volume}{29}, \bibinfo{number}{2} (\bibinfo{year}{2005}), \bibinfo{pages}{472--484}.
\newblock


\bibitem[Li et~al\mbox{.}(2022)]%
        {li2022billion}
\bibfield{author}{\bibinfo{person}{Kunpeng Li}, \bibinfo{person}{Guangcui Shao}, \bibinfo{person}{Naijun Yang}, \bibinfo{person}{Xiao Fang}, {and} \bibinfo{person}{Yang Song}.} \bibinfo{year}{2022}\natexlab{}.
\newblock \showarticletitle{Billion-user customer lifetime value prediction: an industrial-scale solution from Kuaishou}. In \bibinfo{booktitle}{\emph{CIKM}}. \bibinfo{pages}{3243--3251}.
\newblock


\bibitem[Li et~al\mbox{.}(2025)]%
        {li2025g}
\bibfield{author}{\bibinfo{person}{Yuhan Li}, \bibinfo{person}{Xinni Zhang}, \bibinfo{person}{Linhao Luo}, \bibinfo{person}{Heng Chang}, \bibinfo{person}{Yuxiang Ren}, \bibinfo{person}{Irwin King}, {and} \bibinfo{person}{Jia Li}.} \bibinfo{year}{2025}\natexlab{}.
\newblock \showarticletitle{G-Refer: Graph Retrieval-Augmented Large Language Model for Explainable Recommendation}. In \bibinfo{booktitle}{\emph{Proceedings of the ACM on Web Conference 2025}}. \bibinfo{pages}{240--251}.
\newblock


\bibitem[Lin et~al\mbox{.}(2019)]%
        {lin2019pareto}
\bibfield{author}{\bibinfo{person}{Xiao Lin}, \bibinfo{person}{Hongjie Chen}, \bibinfo{person}{Changhua Pei}, \bibinfo{person}{Fei Sun}, \bibinfo{person}{Xuanji Xiao}, \bibinfo{person}{Hanxiao Sun}, \bibinfo{person}{Yongfeng Zhang}, \bibinfo{person}{Wenwu Ou}, {and} \bibinfo{person}{Peng Jiang}.} \bibinfo{year}{2019}\natexlab{}.
\newblock \showarticletitle{A pareto-efficient algorithm for multiple objective optimization in e-commerce recommendation}. In \bibinfo{booktitle}{\emph{Proceedings of the 13th ACM Conference on recommender systems}}. \bibinfo{pages}{20--28}.
\newblock


\bibitem[Liu et~al\mbox{.}(2020)]%
        {liu2020modelling}
\bibfield{author}{\bibinfo{person}{Yang Liu}, \bibinfo{person}{Liang Chen}, \bibinfo{person}{Xiangnan He}, \bibinfo{person}{Jiaying Peng}, \bibinfo{person}{Zibin Zheng}, {and} \bibinfo{person}{Jie Tang}.} \bibinfo{year}{2020}\natexlab{}.
\newblock \showarticletitle{Modelling high-order social relations for item recommendation}.
\newblock \bibinfo{journal}{\emph{TKDE}} \bibinfo{volume}{34}, \bibinfo{number}{9} (\bibinfo{year}{2020}), \bibinfo{pages}{4385--4397}.
\newblock


\bibitem[Lomurno et~al\mbox{.}(2021)]%
        {lomurno2021pareto}
\bibfield{author}{\bibinfo{person}{Eugenio Lomurno}, \bibinfo{person}{Stefano Samele}, \bibinfo{person}{Matteo Matteucci}, {and} \bibinfo{person}{Danilo Ardagna}.} \bibinfo{year}{2021}\natexlab{}.
\newblock \showarticletitle{Pareto-optimal progressive neural architecture search}. In \bibinfo{booktitle}{\emph{Proceedings of the Genetic and Evolutionary Computation Conference Companion}}. \bibinfo{pages}{1726--1734}.
\newblock


\bibitem[Mahapatra and Rajan(2020)]%
        {mahapatra2020multi}
\bibfield{author}{\bibinfo{person}{Debabrata Mahapatra} {and} \bibinfo{person}{Vaibhav Rajan}.} \bibinfo{year}{2020}\natexlab{}.
\newblock \showarticletitle{Multi-task learning with user preferences: Gradient descent with controlled ascent in pareto optimization}. In \bibinfo{booktitle}{\emph{ICML}}. PMLR, \bibinfo{pages}{6597--6607}.
\newblock


\bibitem[Min et~al\mbox{.}(2023)]%
        {min2023scenario}
\bibfield{author}{\bibinfo{person}{Erxue Min}, \bibinfo{person}{Da Luo}, \bibinfo{person}{Kangyi Lin}, \bibinfo{person}{Chunzhen Huang}, {and} \bibinfo{person}{Yang Liu}.} \bibinfo{year}{2023}\natexlab{}.
\newblock \showarticletitle{Scenario-adaptive feature interaction for click-through rate prediction}. In \bibinfo{booktitle}{\emph{KDD}}. \bibinfo{pages}{4661--4672}.
\newblock


\bibitem[Norris(1998)]%
        {norris1998markov}
\bibfield{author}{\bibinfo{person}{James~R Norris}.} \bibinfo{year}{1998}\natexlab{}.
\newblock \bibinfo{booktitle}{\emph{Markov chains}}.
\newblock Number~2. \bibinfo{publisher}{Cambridge university press}.
\newblock


\bibitem[Pan et~al\mbox{.}(2018)]%
        {pan2018field}
\bibfield{author}{\bibinfo{person}{Junwei Pan}, \bibinfo{person}{Jian Xu}, \bibinfo{person}{Alfonso~Lobos Ruiz}, \bibinfo{person}{Wenliang Zhao}, \bibinfo{person}{Shengjun Pan}, \bibinfo{person}{Yu Sun}, {and} \bibinfo{person}{Quan Lu}.} \bibinfo{year}{2018}\natexlab{}.
\newblock \showarticletitle{Field-weighted factorization machines for click-through rate prediction in display advertising}. In \bibinfo{booktitle}{\emph{WWW}}. \bibinfo{pages}{1349--1357}.
\newblock


\bibitem[Sheng et~al\mbox{.}(2021)]%
        {sheng2021one}
\bibfield{author}{\bibinfo{person}{Xiang-Rong Sheng}, \bibinfo{person}{Liqin Zhao}, \bibinfo{person}{Guorui Zhou}, \bibinfo{person}{Xinyao Ding}, \bibinfo{person}{Binding Dai}, \bibinfo{person}{Qiang Luo}, \bibinfo{person}{Siran Yang}, \bibinfo{person}{Jingshan Lv}, \bibinfo{person}{Chi Zhang}, \bibinfo{person}{Hongbo Deng}, {et~al\mbox{.}}} \bibinfo{year}{2021}\natexlab{}.
\newblock \showarticletitle{One model to serve all: Star topology adaptive recommender for multi-domain ctr prediction}. In \bibinfo{booktitle}{\emph{CIKM}}. \bibinfo{pages}{4104--4113}.
\newblock


\bibitem[Singh et~al\mbox{.}(2018)]%
        {singh2018customer}
\bibfield{author}{\bibinfo{person}{Lavneet Singh}, \bibinfo{person}{Nancy Kaur}, {and} \bibinfo{person}{Girija Chetty}.} \bibinfo{year}{2018}\natexlab{}.
\newblock \showarticletitle{Customer Life Time Value Model Framework Using Gradient Boost Trees with RANSAC Response Regularization}. \bibinfo{pages}{1--8}.
\newblock
\href{https://doi.org/10.1109/IJCNN.2018.8489710}{doi:\nolinkurl{10.1109/IJCNN.2018.8489710}}


\bibitem[Su et~al\mbox{.}(2023)]%
        {su2023cross}
\bibfield{author}{\bibinfo{person}{Hongzu Su}, \bibinfo{person}{Zhekai Du}, \bibinfo{person}{Jingjing Li}, \bibinfo{person}{Lei Zhu}, {and} \bibinfo{person}{Ke Lu}.} \bibinfo{year}{2023}\natexlab{}.
\newblock \showarticletitle{Cross-domain adaptative learning for online advertisement customer lifetime value prediction}. In \bibinfo{booktitle}{\emph{AAAI}}, Vol.~\bibinfo{volume}{37}. \bibinfo{pages}{4605--4613}.
\newblock


\bibitem[Sun et~al\mbox{.}(2019)]%
        {sun2019multi}
\bibfield{author}{\bibinfo{person}{Jianing Sun}, \bibinfo{person}{Yingxue Zhang}, \bibinfo{person}{Chen Ma}, \bibinfo{person}{Mark Coates}, \bibinfo{person}{Huifeng Guo}, \bibinfo{person}{Ruiming Tang}, {and} \bibinfo{person}{Xiuqiang He}.} \bibinfo{year}{2019}\natexlab{}.
\newblock \showarticletitle{Multi-graph convolution collaborative filtering}. In \bibinfo{booktitle}{\emph{ICDM}}. IEEE, \bibinfo{pages}{1306--1311}.
\newblock


\bibitem[Tian et~al\mbox{.}(2023)]%
        {tian2023heterogeneous}
\bibfield{author}{\bibinfo{person}{Yijun Tian}, \bibinfo{person}{Kaiwen Dong}, \bibinfo{person}{Chunhui Zhang}, \bibinfo{person}{Chuxu Zhang}, {and} \bibinfo{person}{Nitesh~V Chawla}.} \bibinfo{year}{2023}\natexlab{}.
\newblock \showarticletitle{Heterogeneous graph masked autoencoders}. In \bibinfo{booktitle}{\emph{AAAI}}, Vol.~\bibinfo{volume}{37}. \bibinfo{pages}{9997--10005}.
\newblock


\bibitem[Vanderveld et~al\mbox{.}(2016)]%
        {vanderveld2016engagement}
\bibfield{author}{\bibinfo{person}{Ali Vanderveld}, \bibinfo{person}{Addhyan Pandey}, \bibinfo{person}{Angela Han}, {and} \bibinfo{person}{Rajesh Parekh}.} \bibinfo{year}{2016}\natexlab{}.
\newblock \showarticletitle{An engagement-based customer lifetime value system for e-commerce}. In \bibinfo{booktitle}{\emph{KDD}}. \bibinfo{pages}{293--302}.
\newblock


\bibitem[Wang et~al\mbox{.}(2024)]%
        {wang2024adsnet}
\bibfield{author}{\bibinfo{person}{Ruize Wang}, \bibinfo{person}{Hui Xu}, \bibinfo{person}{Ying Cheng}, \bibinfo{person}{Qi He}, \bibinfo{person}{Xing Zhou}, \bibinfo{person}{Rui Feng}, \bibinfo{person}{Wei Xu}, \bibinfo{person}{Lei Huang}, {and} \bibinfo{person}{Jie Jiang}.} \bibinfo{year}{2024}\natexlab{}.
\newblock \showarticletitle{ADSNet: Cross-Domain LTV Prediction with an Adaptive Siamese Network in Advertising}. In \bibinfo{booktitle}{\emph{KDD}}. \bibinfo{pages}{5872--5881}.
\newblock


\bibitem[Wang et~al\mbox{.}(2019a)]%
        {wang2019neural}
\bibfield{author}{\bibinfo{person}{Xiang Wang}, \bibinfo{person}{Xiangnan He}, \bibinfo{person}{Meng Wang}, \bibinfo{person}{Fuli Feng}, {and} \bibinfo{person}{Tat-Seng Chua}.} \bibinfo{year}{2019}\natexlab{a}.
\newblock \showarticletitle{Neural graph collaborative filtering}. In \bibinfo{booktitle}{\emph{Proceedings of the 42nd international ACM SIGIR conference on Research and development in Information Retrieval}}. \bibinfo{pages}{165--174}.
\newblock


\bibitem[Wang et~al\mbox{.}(2019b)]%
        {wang2019deep}
\bibfield{author}{\bibinfo{person}{Xiaojing Wang}, \bibinfo{person}{Tianqi Liu}, {and} \bibinfo{person}{Jingang Miao}.} \bibinfo{year}{2019}\natexlab{b}.
\newblock \showarticletitle{A deep probabilistic model for customer lifetime value prediction}.
\newblock \bibinfo{journal}{\emph{arXiv preprint arXiv:1912.07753}} (\bibinfo{year}{2019}).
\newblock


\bibitem[Weng et~al\mbox{.}(2024)]%
        {weng2024optdist}
\bibfield{author}{\bibinfo{person}{Yunpeng Weng}, \bibinfo{person}{Xing Tang}, \bibinfo{person}{Zhenhao Xu}, \bibinfo{person}{Fuyuan Lyu}, \bibinfo{person}{Dugang Liu}, \bibinfo{person}{Zexu Sun}, {and} \bibinfo{person}{Xiuqiang He}.} \bibinfo{year}{2024}\natexlab{}.
\newblock \showarticletitle{OptDist: Learning Optimal Distribution for Customer Lifetime Value Prediction}. In \bibinfo{booktitle}{\emph{CIKM}}. \bibinfo{pages}{2523--2533}.
\newblock


\bibitem[Wu et~al\mbox{.}(2022)]%
        {wu2022multi}
\bibfield{author}{\bibinfo{person}{Haolun Wu}, \bibinfo{person}{Chen Ma}, \bibinfo{person}{Bhaskar Mitra}, \bibinfo{person}{Fernando Diaz}, {and} \bibinfo{person}{Xue Liu}.} \bibinfo{year}{2022}\natexlab{}.
\newblock \showarticletitle{A multi-objective optimization framework for multi-stakeholder fairness-aware recommendation}.
\newblock \bibinfo{journal}{\emph{ACM Transactions on Information Systems}} \bibinfo{volume}{41}, \bibinfo{number}{2} (\bibinfo{year}{2022}), \bibinfo{pages}{1--29}.
\newblock


\bibitem[Xie et~al\mbox{.}(2021)]%
        {xie2021personalized}
\bibfield{author}{\bibinfo{person}{Ruobing Xie}, \bibinfo{person}{Yanlei Liu}, \bibinfo{person}{Shaoliang Zhang}, \bibinfo{person}{Rui Wang}, \bibinfo{person}{Feng Xia}, {and} \bibinfo{person}{Leyu Lin}.} \bibinfo{year}{2021}\natexlab{}.
\newblock \showarticletitle{Personalized approximate pareto-efficient recommendation}. In \bibinfo{booktitle}{\emph{WWW}}. \bibinfo{pages}{3839--3849}.
\newblock


\bibitem[Xing et~al\mbox{.}(2021)]%
        {xing2021learning}
\bibfield{author}{\bibinfo{person}{Mingzhe Xing}, \bibinfo{person}{Shuqing Bian}, \bibinfo{person}{Wayne~Xin Zhao}, \bibinfo{person}{Zhen Xiao}, \bibinfo{person}{Xinji Luo}, \bibinfo{person}{Cunxiang Yin}, \bibinfo{person}{Jing Cai}, {and} \bibinfo{person}{Yancheng He}.} \bibinfo{year}{2021}\natexlab{}.
\newblock \showarticletitle{Learning reliable user representations from volatile and sparse data to accurately predict customer lifetime value}. In \bibinfo{booktitle}{\emph{KDD}}. \bibinfo{pages}{3806--3816}.
\newblock


\bibitem[Yang et~al\mbox{.}(2023)]%
        {yang2023feature}
\bibfield{author}{\bibinfo{person}{Xuejiao Yang}, \bibinfo{person}{Binfeng Jia}, \bibinfo{person}{Shuangyang Wang}, {and} \bibinfo{person}{Shijie Zhang}.} \bibinfo{year}{2023}\natexlab{}.
\newblock \showarticletitle{Feature Missing-aware Routing-and-Fusion Network for Customer Lifetime Value Prediction in Advertising}. In \bibinfo{booktitle}{\emph{WSDM}}. \bibinfo{pages}{1030--1038}.
\newblock


\bibitem[Yang et~al\mbox{.}(2022)]%
        {yang2022adasparse}
\bibfield{author}{\bibinfo{person}{Xuanhua Yang}, \bibinfo{person}{Xiaoyu Peng}, \bibinfo{person}{Penghui Wei}, \bibinfo{person}{Shaoguo Liu}, \bibinfo{person}{Liang Wang}, {and} \bibinfo{person}{Bo Zheng}.} \bibinfo{year}{2022}\natexlab{}.
\newblock \showarticletitle{Adasparse: Learning adaptively sparse structures for multi-domain click-through rate prediction}. In \bibinfo{booktitle}{\emph{CIKM}}. \bibinfo{pages}{4635--4639}.
\newblock


\bibitem[Yun et~al\mbox{.}(2023)]%
        {10.1145/3580305.3599871}
\bibfield{author}{\bibinfo{person}{Junwoo Yun}, \bibinfo{person}{Wonryeol Kwak}, {and} \bibinfo{person}{Joohyun Kim}.} \bibinfo{year}{2023}\natexlab{}.
\newblock \showarticletitle{Multi Datasource LTV User Representation (MDLUR)}. In \bibinfo{booktitle}{\emph{KDD}}. \bibinfo{address}{New York, NY, USA}, \bibinfo{pages}{5500–5508}.
\newblock
\showISBNx{9798400701030}
\href{https://doi.org/10.1145/3580305.3599871}{doi:\nolinkurl{10.1145/3580305.3599871}}


\bibitem[Zhang et~al\mbox{.}(2024)]%
        {zhang2024elastst}
\bibfield{author}{\bibinfo{person}{Jiawen Zhang}, \bibinfo{person}{Shun Zheng}, \bibinfo{person}{Xumeng Wen}, \bibinfo{person}{Xiaofang Zhou}, \bibinfo{person}{Jiang Bian}, {and} \bibinfo{person}{Jia Li}.} \bibinfo{year}{2024}\natexlab{}.
\newblock \showarticletitle{ElasTST: Towards Robust Varied-Horizon Forecasting with Elastic Time-Series Transformer}.
\newblock \bibinfo{journal}{\emph{arXiv preprint arXiv:2411.01842}} (\bibinfo{year}{2024}).
\newblock


\bibitem[Zhang et~al\mbox{.}(2021)]%
        {zhang2021wide}
\bibfield{author}{\bibinfo{person}{Mingwang Zhang}, \bibinfo{person}{Kun Huang}, {and} \bibinfo{person}{Yanli Lv}.} \bibinfo{year}{2021}\natexlab{}.
\newblock \showarticletitle{A wide neighborhood arc-search interior-point algorithm for convex quadratic programming with box constraints and linear constraints}.
\newblock \bibinfo{journal}{\emph{Optimization and Engineering}} (\bibinfo{year}{2021}), \bibinfo{pages}{1--21}.
\newblock


\bibitem[Zhang et~al\mbox{.}(2023)]%
        {zhang2023out}
\bibfield{author}{\bibinfo{person}{Shijie Zhang}, \bibinfo{person}{Xin Yan}, \bibinfo{person}{Xuejiao Yang}, \bibinfo{person}{Binfeng Jia}, {and} \bibinfo{person}{Shuangyang Wang}.} \bibinfo{year}{2023}\natexlab{}.
\newblock \showarticletitle{Out of the Box Thinking: Improving Customer Lifetime Value Modelling via Expert Routing and Game Whale Detection}. In \bibinfo{booktitle}{\emph{CIKM}}. \bibinfo{pages}{3206--3215}.
\newblock


\bibitem[Zhou et~al\mbox{.}(2023)]%
        {zhou2023temporal}
\bibfield{author}{\bibinfo{person}{Haolin Zhou}, \bibinfo{person}{Junwei Pan}, \bibinfo{person}{Xinyi Zhou}, \bibinfo{person}{Xihua Chen}, \bibinfo{person}{Jie Jiang}, \bibinfo{person}{Xiaofeng Gao}, {and} \bibinfo{person}{Guihai Chen}.} \bibinfo{year}{2023}\natexlab{}.
\newblock \showarticletitle{Temporal interest network for click-through rate prediction}.
\newblock \bibinfo{journal}{\emph{arXiv preprint arXiv:2308.08487}} (\bibinfo{year}{2023}).
\newblock


\bibitem[Zhou et~al\mbox{.}(2021)]%
        {zhou2021informer}
\bibfield{author}{\bibinfo{person}{Haoyi Zhou}, \bibinfo{person}{Shanghang Zhang}, \bibinfo{person}{Jieqi Peng}, \bibinfo{person}{Shuai Zhang}, \bibinfo{person}{Jianxin Li}, \bibinfo{person}{Hui Xiong}, {and} \bibinfo{person}{Wancai Zhang}.} \bibinfo{year}{2021}\natexlab{}.
\newblock \showarticletitle{Informer: Beyond efficient transformer for long sequence time-series forecasting}. In \bibinfo{booktitle}{\emph{AAAI}}, Vol.~\bibinfo{volume}{35}. \bibinfo{pages}{11106--11115}.
\newblock


\bibitem[Zhou et~al\mbox{.}(2024a)]%
        {zhou2024a}
\bibfield{author}{\bibinfo{person}{Zhiyuan Zhou}, \bibinfo{person}{Li Lin}, \bibinfo{person}{Hai Wang}, \bibinfo{person}{Xiaolei Zhou}, \bibinfo{person}{Wei Gong}, {and} \bibinfo{person}{Shuai Wang}.} \bibinfo{year}{2024}\natexlab{a}.
\newblock \showarticletitle{A Cross Domain Method for Customer Lifetime Value Prediction in Supply Chain Platform}. In \bibinfo{booktitle}{\emph{The Web Conference 2024}}.
\newblock
\urldef\tempurl%
\url{https://openreview.net/forum?id=NwpvPL66Ps}
\showURL{%
\tempurl}


\bibitem[Zhou et~al\mbox{.}(2024b)]%
        {zhou2024cross}
\bibfield{author}{\bibinfo{person}{Zhiyuan Zhou}, \bibinfo{person}{Li Lin}, \bibinfo{person}{Hai Wang}, \bibinfo{person}{Xiaolei Zhou}, \bibinfo{person}{Gong Wei}, {and} \bibinfo{person}{Shuai Wang}.} \bibinfo{year}{2024}\natexlab{b}.
\newblock \showarticletitle{A Cross Domain Method for Customer Lifetime Value Prediction in Supply Chain Platform}. In \bibinfo{booktitle}{\emph{WWW}}. \bibinfo{pages}{4037--4046}.
\newblock


\bibitem[Zhou et~al\mbox{.}(2024c)]%
        {DBLP:conf/iclr/Zhou0Z024}
\bibfield{author}{\bibinfo{person}{Zhipeng Zhou}, \bibinfo{person}{Liu Liu}, \bibinfo{person}{Peilin Zhao}, {and} \bibinfo{person}{Wei Gong}.} \bibinfo{year}{2024}\natexlab{c}.
\newblock \showarticletitle{Pareto Deep Long-Tailed Recognition: {A} Conflict-Averse Solution}. In \bibinfo{booktitle}{\emph{ICLR}}. \bibinfo{publisher}{OpenReview.net}.
\newblock


\end{thebibliography}
